\documentclass{article}

\usepackage{microtype}
\usepackage{graphicx}
\usepackage{booktabs} % for professional tables
\usepackage{multirow}
\usepackage{amsfonts}
\usepackage{amssymb}
\usepackage{amsmath}
\usepackage{siunitx}
\usepackage{array, makecell} \usepackage{wrapfig}

\usepackage{comment}
\usepackage{hyperref}

\newtheorem{definition}{Definition}[section]

\usepackage[accepted]{icml2021}

\icmltitlerunning{Geometric Component Analysis}

\begin{document}

\twocolumn[
\icmltitle{GeomCA: Geometric Evaluation of Data Representations }



\icmlsetsymbol{equal}{*}

\begin{icmlauthorlist}
\icmlauthor{Petra Poklukar}{to}
\icmlauthor{Anastasia Varava}{to}
\icmlauthor{Danica Kragic}{to}
\end{icmlauthorlist}

\icmlaffiliation{to}{KTH Royal Institute of Technology, Stockholm, Sweden}
\icmlcorrespondingauthor{Petra Poklukar}{poklukar@kth.se}

\icmlkeywords{Machine Learning, ICML}

\vskip 0.3in
]



\printAffiliationsAndNotice{}  % leave blank if no need to mention equal contribution

\begin{abstract}
Evaluating the quality of learned representations without relying on a downstream task remains one of the challenges in representation learning.  In this work, we present Geometric Component Analysis (GeomCA) algorithm that evaluates representation spaces based on their geometric and topological properties. 
GeomCA can be applied to representations of any dimension, independently of the model that generated them. We demonstrate its applicability by analyzing representations obtained from a variety of scenarios, such as contrastive learning models, generative models and supervised learning models. 
\end{abstract}

\section{Introduction}
Efficient data representations have been shown to improve machine learning models in numerous domains such as supervised and transfer learning~\cite{transfer_repr_neurips, wang2020unsupervised}, density estimation \cite{kirichenko2020normalizing}, reinforcement learning~\cite{rlrepr-ghosh20b}, to name a few. 
Significant progress has been made on learning representations with different structures, for example disentangled \cite{pfau2020disentangling_subspace, kim_factorvae}, in the form of a specific manifold or curvature \cite{arvanitidis2018latent, arvanitidis2020geometrically,  moor2020topological, r2020witness}, or with particular similarity as learned by contrastive learning algorithms \cite{le2020contrastive}.
%a particular structure such as disentangled \cite{pfau2020disentangling_subspace, kim_factorvae}, with a particular geometric properties such as manifold structure or curvature \cite{arvanitidis2018latent, arvanitidis2020geometrically,  moor2020topological, r2020witness} or with particular similarity structure, for example learned by contrastive learning algorithms \cite{le2020contrastive}. %These techniques mostly rely on unsupervised training approaches by either deploying self-supervised models or Variational Autoencoders CITE. [CITE LSR] [Mnetion the dim reduction!]
In general, data representations are desirable not only due to their low dimensionality but also because they enable measuring meaningful distances, which is especially critical in noisy data or visual data such as images and videos.
%or lack of meaningful notion of distance in raw data, which is especially critical in visual data such as images and videos.

%various hindrances of the raw data space. These include presence of noise or lack of meaningful notion of distance which is especially critical in visual data such as images and videos.

%Learning descriptive representations is especially important for visual data, such as images and videos, since these domains are lacking reliable notion of distance. In particular, metrics operating on raw pixel values are not robust to translations and color transformations. Therefore, when working with visual data, the most common way to extract high-level features is to either use pre-trained models trained on generic classification datasets or extract representations using the above mentioned unsupervised approaches. Working with raw data is problematic even for other data formats, such as text or signal data, due to the particular sequential semantic information or presence of noise.

The quality of learned representations is typically determined by their performance on a specific downstream task. 
%they yield when given as inputs to a specific downstream task.
%evaluated using downstream tasks, where representations are assumed to be efficient if we obtain high performance on the given downstream task. 
%For example, the level of both disentanglement and contrastivness (separability) the model is able to achieve are evaluated using  classification models. In the former case, the task is to predict the label of the underlying factor of variation present in the data, while in the latter the label of...
For example, disentanglement is determined by the accuracy of classifiers trained to predict the underlying factors of variation present in the dataset \cite{higgins2016beta, kim_factorvae, locatello2019challenging}. %Similarly, contrastiveness is evaluated using a classifier trained on data labels. 
In reinforcement learning and robotics, usefulness of representations is evaluated on the performance of the policy \cite{ghadirzadeh2020data, rl_representetions}  and the robotics task \cite{lippi2020latent}, respectively. %In latent variable models, evaluation is commonly performed by visual inspection of generated samples. 
However, such evaluation favors representations that are tied to the downstream task, making them 
difficult to be generalized across variety of tasks. %For example, in reinforcement learning, even minor changes in the policy make same representations difficult to be reused. 
%To allow generalization, representations should express the overall structure of the underlying data manifold.
%However, this approach does not allow to create general representations that are useful for a variety of tasks. To be reusable for different purposes, representations should reflect the overall structure of the underlying data manifold.

%However, such evaluation approaches provide little insight into the actual structure of learned representations, making it difficult to investigate potential failure cases. {\color{blue} Furthermore, this approach does not allow to create general representations that can be useful for a variety of tasks.} %To perform such diagnosis, geometric information of data representations should be taken into account. %during evaluation.

A more general way to evaluate representations is to analyze how well their \emph{global structure}, i.e., their geometric and topological properties, reflect the underlying structure of the data manifold. % space. %To be reusable for different purposes, a representation should reflect the overall structure of the underlying distribution, i.e., to have a meaningful notion of distance between points, encode similar datapoints close to each other, and separate the dissimilar ones~\cite{jang2019need}. }
This direction has been recently explored in the context of Generative Adversarial Networks (GANs)~\cite{gans} where it is challenging to define an appropriate downstream task for evaluation. Recently proposed methods, such as Improved Precision and Recall (IPR)~\cite{ipr} and Geometry Score (GS)~\cite{GSkhrulkov18a}, have shown success in detecting failure cases of GANs but provide little insight into the actual \textit{structure} of learned representations, %making it difficult to investigate local areas of representations space where the potential failures arise.
thus hindering further investigation of local areas of the representation space where the potential failures arise.

\begin{figure}[ht]
\vskip 0.2in
\begin{center}
\centerline{\includegraphics[width=0.9\columnwidth]{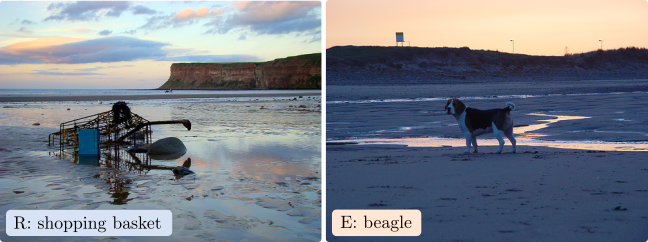}}
\caption{Example of two images of different class label in the ImageNet dataset belonging to the same connected component in the VGG16 representation space.}
\label{fig:exp:vgg16-front}
\end{center}
\vskip -0.1in
\end{figure}

%{\color{blue} In this work, we present a method for evaluating  the structure of both the entire representation space and any of its connected components. Given a set of representations $E$ that we wish to evaluate, our method analyzes how well it reflects the structure of the reference set of representations $R$ that are assumed to reflect the underlying data manifold.Our method, called Geometric Component Analysis (GeomCA), }

In this work, we present %Geometric Component Analysis (GeomCA) algorithm 
a method for evaluating the structure of both the entire representation space and any of its connected components. We achieve this by comparing two discrete sets representing the true data manifold: the reference representation set $R$ and the evaluation representation set $E$. Depending on the application, $R$ and $E$ can consist of raw data points or their features %encodings 
obtained from a neural network. 
Intuitively, if the structure of evaluated representations $E$ is well aligned with the structure of the reference representations $R$, then $E$ well represents the underlying data manifold.

%Our method, called Geometric Component Analysis (GeomCA),  analyzes the degree of their the alignment between $R$ and $E$ not only on a \textit{global} level, as done by other closely related methods such as GS and IPR, but also on a \textit{local} level by analyzing the location and correspondence between the connected components of $R$ and $E$.

Our method, called Geometric Component Analysis (GeomCA), uses graphs to extract the structure of $R$ and $E$, %using graphs, 
and analyzes the degree of their alignment. In contrast to other closely related methods such GS and IPR, we analyze the alignment not only on a \textit{global} level,
%not only on a \textit{global} level, as done by other closely related methods such as GS and IPR, 
but also on a \textit{local} level by analyzing the location and correspondence between the connected components of $R$ and $E$. We demonstrate
that GeomCA can detect outliers and connected components in $E$ that are not present in $R$, as well as identify the coordinates of individual points from any component, thus enabling their visualization.% of the data from a specific component or an outlier.

\comment{ In this work, we consider two discrete sets of points representing the true data manifold: the reference representation set $R$ and the evaluated representation set $E$. Depending on the application, $R$ and $E$ can consist of datapoints or their feature encodings obtained from a neural network.
We present a method for evaluating  the structure of both the entire representation set $E$ and any of its connected components.
%evaluating the representation space structure. 
In contrast to other closely related methods, such GS and IPR, our method complements the \emph{global} analysis of the {\color{blue} learned} data manifold with \emph{local} information about the location of its connected components, and the degree of their correspondence to the respective connected components of the {\color{blue} reference} %true 
data representation set.

%In contrast to other closely related methods, such GS and IPR that only analyze the \emph{global} structure of the representation space, out method complements it with \emph{local} information about the location of the connected components, their size, and the degree of their correspondence to the respective connected components of the true data space.

Our method, called Geometric Component Analysis (GeomCA), analyzes how well representations of interest $E$ reflect the structure of the reference representations $R$.
%evaluates the structure of the representation space by analyzing the alignment between representations of interest, denoted by $E$
%----evaluates how  well the representations %to be evaluated, of interest, denoted by $E$, reflect the structure of the reference representations $R$. 
%Intuitively, if both of the sets reflect the true diversity of the data, then the connected components of the underlying true data manifold contain points from both $R$ and $E$.% which are also well aligned.  [MANIFOLD CONFUSION]
Intuitively, if the structure of the evaluated representation $E$ is well aligned with the structure of the reference $R$, then $E$ represents the underlying data manifold. We demonstrate
that GeomCA can both detect outliers and connected components in $E$ that are not present in $R$, and identify the coordinates of any individual point from any component, making it possible to visualize the data from a specific component or an outlier.
}
%several advantages compared to related approaches:
%that our method has a number of advantages comparing to related approaches: % such as Geometry Score: in particular, 
%it allows us to detect outliers and connected components in $E$ that are not present in $R$; furthermore, it allows to identify the coordinates of any individual point from any component, making it possible to visualize the data from a specific component or an outlier.

We apply GeomCA to different practical setups. First, we consider a contrastive learning scenario and %apply GeomCA to 
evaluate the structural similarity between the encodings belonging to different classes of the training and validation datasets (Section~\ref{sec:exp:multiview-box}).
Second, %we use GeomCA to 
we evaluate generative models by comparing the connected components of the training and generated datasets (Section~\ref{sec:exp:gan}). Finally, we apply GeomCA to investigate if features extracted by a supervised model are separated according to their respective classes (Section~\ref{sec:exp:vgg16}).   %whether features extracted by a supervised model achieve class separation (Section~\ref{sec:exp:vgg16}). 
For instance, Figure~\ref{fig:exp:vgg16-front} shows two images belonging to different classes in the ImageNet dataset~\cite{deng2009imagenet} that are close to each other in the feature space of VGG16 \cite{vgg16}. 

In summary, our contributions are: \textit{(i)} we present GeomCA for assessing the quality of data representations by leveraging their geometric and topological properties (Sections \ref{sec:method}, \ref{sec:implementation-details}), and \textit{(ii)} we experimentally demonstrate valuable insights provided by GeomCA on representations obtained from various models and scenarios (Sections~\ref{sec:exp:multiview-box}, \ref{sec:exp:gan}, \ref{sec:exp:vgg16}).

\section{Method} \label{sec:method}

In this section, we introduce the proposed GeomCA algorithm. We present its intuitive idea in Section~\ref{sec:method:idea}, its technical details in Section~\ref{sec:method:alg} and its improvements over the existing closely related methods in Section~\ref{sec:method:comparisson}.

%In this section, we present the intuitive idea of GeomCA (Section~\ref{sec:method:idea}), its technical details (Section~\ref{sec:method:alg}) and its improvements over the existing closely related methods (Section~\ref{sec:method:comparisson}).

\subsection{Intuitive Idea} \label{sec:method:idea}
The basic idea of the GeomCA algorithm is to compare the global properties (topology) and local properties (geometry) of two sets of representations,
$R$ and $E$, representing the underlying true data manifold $\mathcal{M}$. We say that $E$ is a good representation of $\mathcal{M}$ if it is well aligned with the reference representation $R$. We aim to detect areas where $R$ and $E$ are coherent and quantify %the degree of 
their alignment, as well as detect isolated individual representations (outliers) or groups of points  
%representations consisting of points 
from only one of the sets $R$ or $E$. In summary, we wish to answer the following two questions: 
\begin{enumerate}
    \item[(Q1)] 
    Do $R$ and $E$ have the same number of connected components, and are their sizes comparable? [Topology]
 
    \item[(Q2)] 
    How much do the connected components of $R$ and $E$ overlap? [Geometry]
\end{enumerate}

\comment{
To answer these questions, we first need to find the connected components of $R$ and $E$. %, compare their size, and quantify their alignment. For this, we build
%These can be extracted from 
For this, we turn to \textit{$\varepsilon$-threshold graphs}, or in short $\varepsilon$-graphs. }

\begin{figure}[ht]
\vskip 0.2in
\begin{center}
\centerline{\includegraphics[width=0.95\columnwidth]{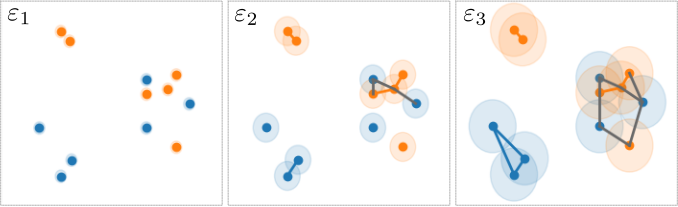}}
\caption{Examples of $\varepsilon$-graphs obtained for $0 < \varepsilon_1 < \varepsilon_2 < \varepsilon_3$ built on the sets $R$ (blue) and $E$ (orange).}
\label{fig:method:vietoris-rips}
\end{center}
\vskip -0.1in
\end{figure}

We find the connected components of $R$ and $E$ by building \textit{$\varepsilon$-threshold graphs}, or in short, $\varepsilon$-graphs. 
In an $\varepsilon$-graph,
%where 
two points are connected by an edge if %their distance is smaller 
they are less than the given threshold $\varepsilon$ apart\footnote{Vertex sets of graph-connected components in an $\varepsilon$-graph are equivalent to clusters obtained from the DBSCAN clustering algorithm~\cite{dbscan}.}.
This allows us to immediately answer (Q1): we build $\varepsilon$-graphs $\mathcal{G}^R, \mathcal{G}^E$ on $R$ and $E$, respectively, and compare the number 
% ADDED
of their connected components as well as their sizes. Examples of $\varepsilon$-graphs for %different choices of the $\varepsilon$ threshold 
$\varepsilon_1 < \varepsilon_2 < \varepsilon_3$ are visualized in Figure \ref{fig:method:vietoris-rips}. In the left panel, $R$ (in blue) and $E$ (in orange) both have $6$ connected components composed of only one point each. When increasing $\varepsilon$, edges among them start emerging (colored with respective color of the set). In the middle panel, $R$ has four connected components of size $1$ and one of size $2$, while $E$ has three connected components of size $1$, $2$ and $3$.

%However, in order 
To answer (Q2), we additionally need to quantify the alignment of these connected components. %$\mathcal{G}^R$ and $\mathcal{G}^E$ 
Intuitively, this can be measured in terms of edges connecting $R$ and $E$. In Figure \ref{fig:method:vietoris-rips}, we visualized such edges in gray color. In the middle panel, the two $R$ components in top right area are well aligned with the largest $E$ component. This area %where $R$ and $E$ align 
increases for a larger $\varepsilon$ shown in the right panel. %For a larger $\varepsilon_3$, the area where $R$ and $E$ align includes all points in the right part.

\begin{figure}[ht]
\vskip 0.2in
\centering
\begin{center}
\centerline{\includegraphics[width=0.9\columnwidth]{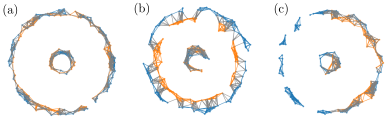}}
\caption{Examples of 2-dimensional points representing the set $R$ (blue) and $E$ (orange) arranged in components of different consistency and quality. %In the table on the right we show the global (top) and local (bottom) scores obtained from GeomCA for each of the arrangement.
}
\label{fig:method:toy-data}
\end{center}
\vskip -0.2in
\end{figure}

We connect points from $R$ and $E$ using the same $\varepsilon$ threshold as in $\mathcal{G}^R$ and $\mathcal{G}^E$. This is equivalent to building an $\varepsilon$-graph $\mathcal{G}^{R \cup E}$ on the union $R \cup E$. %Using $\mathcal{G}^{R \cup E}$, 
In this way, studying the alignment of $\mathcal{G}^R$ and $\mathcal{G}^E$ translates to studying the nature of the connected components of $\mathcal{G}^{R \cup E}$. If $R$ and $E$ are well aligned, all the connected components of $\mathcal{G}^{R \cup E}$ are ``well mixed''. This means, equally \textit{well represented} by points from $R$ and $E$ which are, in turn, also \textit{well geometrically positioned}. For example, in Figure \ref{fig:method:toy-data} (a) and (b), both $\mathcal{G}^{R \cup E}$ graphs have two connected components that are equally well represented by $R$ and $E$. However, in (b), $R$ and $E$ in the outer component are not well mixed but rather concatenated. On contrary, in (c), the two components containing both $R$ and $E$ points are well geometrically aligned but none of the components are equally well represented by $R$ and $E$.

In summary, we evaluate the topological and geometric properties of $R$ and $E$ by investigating the connected components of the $\varepsilon$-graph $\mathcal{G}^{R \cup E}$. We answer (Q1) by analyzing the nature of the vertices in each component, and (Q2) by analyzing the nature of the edges. This intuitive idea is the main driver behind GeomCA, which we rigorously define in the following section.

\subsection{GeomCA Algorithm} \label{sec:method:alg}
Let $X = \{ x_i \}_{i = 1}^{n_X} \subset \mathbb{R}^M$ be a dataset of observations and let $Z = \{ z_i \}_{i = 1}^{n_X} \subset \mathbb{R}^N$ denote their %corresponding 
representations obtained from any model $\mathbb{M}$, i.e., $Z = \mathbb{M}(X)$. In a machine learning setup, 
it is commonly assumed that $N \ll M$, although this
is not necessary for GeomCA to work. Let $R = \{ z_i \}_{i = 1}^{n_R}$ %\subset Z$ 
and $E = \{ z_i \}_{i = 1}^{n_E}$ be two subsets of representations in $Z$ for which we assume that $n_R + n_E \le n_X$ and $R \neq E$. The latter assumption %is needed in order to obtain meaningful insights as it 
eliminates the case where $R$ and $E$ are perfectly aligned. While GeomCA provides most insight into representations when $R \cap E = \emptyset$, a non-empty intersection %$\emptyset \neq R \cap E \subsetneq R \cup E$ 
might be desirable in situations where it is important to investigate deviations from the intersection $R \cap E$.

As intuitively explained in Section \ref{sec:method:idea}, the idea of GeomCA is to analyze the alignment of $R$ and $E$ using $\varepsilon$-threshold graphs defined below.

\begin{definition} \label{def:vr}
An $\varepsilon$-threshold graph, or $\varepsilon$-graph, on the set of points $W$ with respect to the radius $\varepsilon > 0$ is a graph $\mathcal{G}_\varepsilon(W) = (\mathcal{V}, \mathcal{E})$ with vertices $\mathcal{V} = W$ and edges $\mathcal{E} = \{ e_{ij} = (v_i, v_j) \in \mathcal{V} \times \mathcal{V} \, | \, d(v_i, v_j) < \varepsilon \}$.
\end{definition}

We built an $\varepsilon$-graph $\mathcal{G}_\varepsilon(R \cup E)$ on the union $R \cup E$. We discuss the choice of the radius $\varepsilon$ in Section \ref{sec:epsilon_selection}. In the remainder of this section, we will refer to the graph $\mathcal{G}_\varepsilon(R \cup E)$ simply as $\mathcal{G}$, and denote its connected components by $\mathcal{G}_i$ such that $\mathcal{G} := \sqcup_i \mathcal{G}_{i}$. Moreover, we define a restriction of a graph $\mathcal{H}$ to a subset $W$ to be the subgraph $\mathcal{H}^W \subset \mathcal{H}$ with vertex and edge sets restricted to $W$. For example, a connected component $\mathcal{G}_i$ restricted to the set $R$ is a graph $\mathcal{G}_i^R$ obtained by removing all $E$ points from the vertex set as well as all the edges from and to them from the edge set of $\mathcal{G}_i$.
%For example, a connected component $\mathcal{G}_i$ restricted to the set $R$ is a graph $\mathcal{G}_i^R$ obtained by removing all $E$ points as well as all the edges from and to them from the vertex and edge sets of $\mathcal{G}_i$. 
Lastly, we denote by $|\mathcal{H}|_\mathcal{V}$ and $|\mathcal{H}|_\mathcal{E}$ the cardinalities of the vertex set and edge set of a graph $\mathcal{H}$, respectively.

Our algorithm (summarized in Algorithm~\ref{alg:geomCA-alg}) consists of a \emph{local evaluation} and a \emph{global evaluation} phase. 
The former evaluates how well the connected components of $\mathcal{G}$ are represented by $R$ and $E$, while the latter
evaluates the alignment of $R$ and $E$ on the level of the entire graph $\mathcal{G}$. %performs similar analysis on the level of the entire graph $\mathcal{G}$.
We describe each of these phases in detail in the following.% rest of the section.

%we achieve this by constructing a Vietoris-Rips simplicial complex. Since constructing higher-order simplices is computationally expensive, we simplify this construction by considering 1-skeletons of simplicial complexes containing only points and edges between them. Therefore, these form a Vietoris-Rips graph defined below.

\noindent \textbf{Local evaluation} The goal of this phase is to analyze the connected components of %the $\varepsilon$-graph 
$\mathcal{G}$. As mentioned in Section \ref{sec:method:idea}, we study their geometric properties with respect to %representations from 
the sets $R$ and $E$.  In particular, we study their (i) vertex heterogeneity determined by the ratio of representations from $R$ and $E$ contained in them, and (ii) heterogeneity of edges among these vertices. % determining the quality. 
We refer to these geometric properties as \textit{consistency} and \textit{quality} of connected components, respectively, and rigorously define them in the following.

\begin{definition} \label{def:comp-consistency}
Consistency $c$ of a component $\mathcal{G}_i$ is defined as the ratio 
\begin{equation}
    c(\mathcal{G}_i) = 1 - \frac{|\, |\mathcal{G}_i^R|_\mathcal{V} - |\mathcal{G}_i^{E}|_\mathcal{V} \,|}{|\mathcal{G}_i|_\mathcal{V}}.
\end{equation}
\end{definition}

A component $\mathcal{G}_i$ attains high consistency score $c(\mathcal{G}_i)$ if it contains equally many representations from $R$ and $E$, i.e., if $|\mathcal{G}_i^R|_\mathcal{V} \approx |\mathcal{G}_i^E|_\mathcal{V}$. 
% ADDED
We call such component \emph{consistent}. %\textit{vertex-heterogeneous}, or $\mathcal{V}$-heterogeneous. 
On contrary, $c(\mathcal{G}_i)$ is low if $\mathcal{G}_i$ is dominated either by points from $R$ or $E$, %only one of the sets $R$ or $E$, 
in which case $\mathcal{G}_i$ is said to be \emph{inconsistent}. %\textit{$\mathcal{V}$-homogeneous}.
In Figure \ref{fig:method:toy-data}, panels (a) and (b) contain examples of consistent components, while (c) shows inconsistent ones consisting only of points from $R$.
%we shown examples of consistent (panels (a) and (b)) and inconsistent (right) components.
However, as seen in (b), consistency itself is not a sufficient measure as it fails to detect cases where $R$ and $E$ are consistent but not well geometrically positioned. %, which is 
%for evaluating components. In this case, a component is diverse in terms of vertices but not in terms of edges as the 
%the number of edges among representations from $R$ and $E$ is low. 
This is measured by component quality determined by the number of edges among representations from $R$ and $E$ as defined below. 

\begin{definition} \label{def:comp-quality}
Quality of a component $\mathcal{G}_i$ is defined as the ratio
\begin{equation}
    q(\mathcal{G}_i) = 
    \begin{cases}
    1 -  \frac{(| \mathcal{G}_i^{R}|_\mathcal{E} + |\mathcal{G}_i^{E}|_\mathcal{E})}{|\mathcal{G}_i|_\mathcal{E}} & \text{if } |\mathcal{G}_i|_\mathcal{E} \geq 1,\\
    0              & \text{otherwise}.
\end{cases}
\end{equation}
\end{definition}

A component $\mathcal{G}_i$ attains high quality score, if it exhibits good connectivity among representations from $R$ and $E$ it contains, i.e., if both $|\mathcal{G}_i^{R}|_\mathcal{E}$ and $|\mathcal{G}_i^{E}|_\mathcal{E}$ are small.
% ADDED
We call the edges connecting $R$ and $E$ \emph{heterogeneous} and a component with high number of heterogeneous edges to be of \textit{high quality}. On the other hand, if edges in $\mathcal{G}_i$ exist only among points from one of the sets $R$ or $E$, the component achieves low quality score and is said to be of \textit{low quality} and its edges \textit{homogeneous}. 
In Figure~\ref{fig:method:toy-data} (a) and (b), the connected components are consistent but only the ones in (a) are also of high quality (visualised by large number of gray edges). 
% ADDED
On contrary, the components in (c) are inconsistent but the largest component in fact has many heterogeneous edges, thus attaining high quality score.
%The consistency and quality scores of the two largest components of all examples in Figure~\ref{fig:method:toy-data} are shown the accompanying table. 

\begin{algorithm}
\caption{GeomCA} 
\begin{algorithmic}
\REQUIRE sets of representations $R$ and $E$
%\REQUIRE sparsification distance $\delta$
\REQUIRE component consistency thresholds $\eta_c$ 
\REQUIRE component quality threshold $\eta_q$
\REQUIRE Distance threshold $\varepsilon$ 
%\STATE $T', V' \gets \texttt{sparsify}(T, V)$ 
\STATE $\mathcal{G} \gets \texttt{build\_epsilon\_graph}(R, E)$

\STATE \textbf{[Phase: Local evaluation]}
\STATE $\mathcal{C} \gets \texttt{get\_connected\_components}(\mathcal{G})$
\STATE $\mathcal{Q}_{\text{local}} \gets \texttt{zeros(len($\mathcal{C}$), 2)}$
\FOR{$i = 0, \dots, \texttt{len($\mathcal{C}$)}$}
\STATE $\mathcal{G}_i \gets \mathcal{C}[i]$
\STATE compute $c(\mathcal{G}_i)$ as in Definition \ref{def:comp-consistency}
\STATE compute $q(\mathcal{G}_i)$ as in Definition \ref{def:comp-quality}
\STATE $\mathcal{Q}_{\text{local}}[i, :] \gets [c(\mathcal{G}_i), q(\mathcal{G}_i)]$
\ENDFOR
\STATE \textbf{[Phase: Global evaluation]}
\STATE compute $c(\mathcal{G})$ and $q(\mathcal{G})$ as in Definition \ref{def:network_scores}
\STATE compute $\mathcal{P}, \mathcal{R}$ with respect to $\eta_c, \eta_q$ as in Definition \ref{def:geom-pr}
%\STATE compute $q(\mathcal{G})$ as in Definition \ref{def:comp-quality}
\STATE $\mathcal{Q}_{\text{global}} \gets [\mathcal{P}, \mathcal{R},  c(\mathcal{G}), q(\mathcal{G})]$
\end{algorithmic}
\textbf{Return:} $\mathcal{Q}_{\text{global}}$, $\mathcal{Q}_{\text{local}}$
\label{alg:geomCA-alg}
\end{algorithm}

\noindent \textbf{Global evaluation}
The consistency and quality measures can be also used in several ways to obtain global scores over the entire $\varepsilon$-graph. First, we simply generalize Definitions \ref{def:comp-consistency} and \ref{def:comp-quality} to $\mathcal{G}$. 

\begin{definition} \label{def:network_scores}
We define $c(\mathcal{G})$ as \textit{network consistency}, and $q(\mathcal{G})$ as \textit{network quality}. 
\end{definition}

The global network consistency and quality are important measures to detect imbalances between the sets $R$ and $E$. This is especially applicable in large-scale experiments where the sizes of $R$ and $E$ are reduced for computational purposes. We discuss such reduction in Section \ref{sec:gamma} and demonstrate the usefulness of network consistency and quality measures in these situations in Section \ref{sec:exp:gan}. 

Next, we exploit the components of certain consistency and quality to retrieve two more global scores: precision and recall. These are determined by the fraction of points from one set contained in specific components of $\mathcal{G}$.

\begin{definition} \label{def:geom-pr}
Let $\eta_c, \eta_q \in [0, 1]$ be real numbers. Let
\begin{equation}
    \mathcal{S}(\eta_c, \eta_q) = 
    \bigcup_{\substack{ q(\mathcal{G}_i) > \eta_q, \\ c(\mathcal{G}_i) > \eta_c}} \mathcal{G}_i
\end{equation}
be the union of the connected components $\mathcal{G}_i$ with the minimum consistency and quality scores determined by $\eta_c$ and $\eta_q$, respectively. Let $\mathcal{S}(\eta_c, \eta_q)^R$ and $\mathcal{S}(\eta_c, \eta_q)^E$ denote the restrictions of $\mathcal{S}(\eta_c, \eta_q)$ to the sets $R$ and $E$, respectively. We define precision $\mathcal{P}$ and recall $\mathcal{R}$
with respect to $\eta_c, \eta_q$ as 
\begin{equation}
    \mathcal{P} = \frac{|\mathcal{S}^E|_\mathcal{V}}{|\mathcal{G}^E|_\mathcal{V}} \quad \mathcal{R} = \frac{|\mathcal{S}^R|_\mathcal{V}}{|\mathcal{G}^R|_\mathcal{V}},%\quad N = \frac{|\mathcal{S}^T|_\mathcal{V} + |\mathcal{S}^V|_\mathcal{V}}{|\mathcal{G}|_\mathcal{V}}
\end{equation}
respectively, where we omitted the explicit dependency on $\eta_c, \eta_q$ for simplicity. 
\end{definition}

The thresholds $\eta_c$ and $\eta_q$ determine the level of alignment between the sets $R$ and $E$ that we wish to consider, and therefore enable to easily focus our analysis on the connected components of the desired quality and consistency. 
% ADDED
A high value of $\eta_c$ requires the components to be consistent %be equally well represented by the sets $R$ and $E$, 
while a high value of $\eta_q$ additionally requires the components to have large number of heterogeneous edges.
%that the points from $R$ and $E$ in each component are well mixed and consistent. 
The effect of these thresholds is demonstrated in %Supplementary Material. 
Appendix \ref{app:box}.

\subsection{Comparison with Closely Related Methods}\label{sec:method:comparisson}
Our method is in spirit closest to Geometry Score (GS) \cite{GSkhrulkov18a}, and in implementation to Improved Precision and Recall score (IPR) \cite{ipr}. Both of these methods were developed for evaluation of generative models and therefore also use a reference set $R$ consisting of training data, and an evaluation set $E$ consisting of the generated data.
%and evaluation sets, defined as true and generated data, respectively. 
GS first estimates the manifolds described by $R$ and $E$ using Witness complexes, and then
compares their topological properties using persistent homology. The comparison is based on Relative Living Times (RTL) of homology derived from persistence barcodes in a probabilistic form. Because of Witness complexes, GS relies on repetitive subsampling to obtain a stable estimate. GeomCA instead compares topological properties of $R$ and $E$ by analyzing an $\varepsilon$-graph which is equivalent to the $1$-skeleton of a Vietoris-Rips graph at the given threshold $\varepsilon$. In contrast to GS, GeomCA exploits all the samples and does not require subsampling. Compared to GS, GeomCA is much simpler to tune as it depends only one hyperparameter $\varepsilon$ with an intuitive interpretation.

In IPR, the $R$ and $E$ manifolds are approximated using spheres around each point with radius determined by their $k$-nearest neighbours. The hyperparameter $k$ can result in large volumes in sparse areas, which authors resolve with manual pruning. GeomCA could be interpreted as using spheres of fixed radius $\varepsilon$, except that we do not endow the graph with any volume. In order to run IPR, the sets $R$ and $E$ need to have the same size, which is not requirement for neither GS or GeomCA. 

In contrast to both methods, GeomCA not only extracts the connected component but also enables detailed analysis of their structure by investigating the corresponding vertices and edges. Moreover, GeomCA enables flexibility to evaluate components of specific size, consistency or quality. In addition to the detailed local evaluation of the components, our refined metrics also provide insights into the global structure of the representation space.
% ADDED
\section{Implementation Details and Experimental Design} \label{sec:implementation-details}
In this section, we provide additional implementation details as well as an overview of our experiments.

\subsection{Selecting distance threshold $\varepsilon$} \label{sec:epsilon_selection}
The structure of the $\varepsilon$-graph $\mathcal{G}$ depends on the hyperparameter $\varepsilon$ determining the maximum length of its edges. Extracting the true underlying value of $\varepsilon$ is a non-trivial task, especially in higher dimensional representation spaces. If $\varepsilon$ is too small, each point is contained in its own component, while $\varepsilon$ too large connects all the points into one single component. The true $\varepsilon$ that results in the approximation of $\mathcal{M}$ reflecting the correct topology of the space lies between these two extreme choices. A more precise estimate could be determined by topological algorithms, such as persistent homology~\cite{computing_persistence}. However, due to computational and scalability issues of this approach, we instead %either choose a specific value of $\varepsilon$ or estimate it empirically by examining the distances in the reference set $R$. 
% ADDED
resort to a simple practically applicable heuristic and estimate $\varepsilon$ empirically by examining the distances in the reference set $R$. We randomly sample $2k$ representations from $R$ and calculate their pairwise distances $D = \{ d(z_i, z_j) | \, i=1, \dots k, j = k + 1, \dots, 2k \}$. We then set $\varepsilon$ as $p$th percentile of the set $D$ and denote it by $\varepsilon = \varepsilon(p)$.
In our experiments, we chose small $p$ in scenarios where we expect the distances among certain points to have low variance. For example, in contrastive learning, we expect points considered similar to have small distances (Section~\ref{sec:exp:multiview-box}). On the other hand, in high-dimensional representation spaces, we expect the estimated distances to naturally have a larger variance which is why we chose a larger $p$ (Sections~\ref{sec:exp:gan} and \ref{sec:exp:vgg16}). We leave the improvements on the choice of $\varepsilon$ as future work.
% ADDED

\subsection{Reducing the number of representations} \label{sec:gamma}
As seen from Definition \ref{def:vr}, the construction of $\varepsilon$-graph involves calculation of pairwise distances among points in $R \cup E$. Such calculation can become a computational burden when analyzing large sets of representations. One way to reduce the number of representations in $R$ and $E$ without losing the topological information is to perform \textit{geometric sparsification} defined below.

\begin{definition}
A geometric sparsification of a set $W$ with respect to a sparsificaltion distance $\delta > 0$ is a subset $W' \subset W$ such that $d(w_i, w_j) > \delta$ for every $w_i, w_j \in W'$, $i \neq j$.
\end{definition}

The sparsificaiton parameter $\delta$ determines the extend of data reduction, where a larger $\delta$ results in a sparser point cloud. 
We perform geometric sparsification on the sets $R$ and $E$ separately, such that we can construct the $\varepsilon$-graph more efficiently using the obtained sparse sets $R'$ and $E'$. The reason for separate sparsification of $R$ and $E$ is to detect potential differences in their topology. Intuitively, if $R$ and $E$ reflect the structure of the same representation space, then so should the sparsified sets. 
% ADDED

We emphasize that this is an optional pre-processing step added for computational efficiency and can be disregarded if sufficiently powerful hardware is available. Note that the process affects the introduced consistency score $c$ as it reduces the number of points in sets $R$ and $E$ but \textit{does not change their topology} precisely because it takes into account the geometric position of the points. The choice of the sparsification parameter $\delta$ is closely related to the choice of the distance threshold $\varepsilon$, and should be chosen from the interval $[0, \varepsilon]$. Intuitively, if $\delta = \varepsilon$, a component in the $\varepsilon$-graph $\mathcal{G}$ is created only when points from $R$ and $E$ are well mixed. This means that between every pair of points from one set (e.g., $R$) there necessarily needs to exists a point from the other set (e.g., $E$) that is less than $\varepsilon$ apart from both of the points from the first set (e.g., $R$). On the other hand, if $\delta < \varepsilon$, we still allow points from each of the sets to get connected without having a ``witness'' from the other set. %Regardless of the choice of $\delta$, the sparsification process will keep the subset of the most outer points. [IS THIS A RISKY STATEMENT?]. 

\subsection{Experiment Overview}
\label{sec:experiments}
We implemented GeomCA described in Algorithm \ref{alg:geomCA-alg} using GUDHI library~\cite{gudhi:urm} which supports efficient computation of geometric sparsification, and Networkx library~\cite{networx} for building and analyzing $\varepsilon$-graphs. Our code is available on GitHub\footnote{\url{https://github.com/petrapoklukar/GeomCA}}. We applied GeomCA on three different scenarios and evaluated 
\begin{itemize}
    \item similarity of representations obtained from two contrastive learning methods, Siamese \cite{siamese} and SimCLR \cite{simclr},
    
    \item quality and diversity of images generated by a StyleGAN \cite{stylegan}, and
    
    \item separability of representations obtained from a pretrained VGG16 \cite{vgg16} model on ImageNet \cite{deng2009imagenet}.
\end{itemize}

We compared the results with IPR and GS methods using hyperparameters described in %Supplementary Material.  
Appendix \ref{app:experiments}. 
We denote the IPR precision and recall by $I\mathcal{P}$, $I\mathcal{R}$,  respectively, and mark GS scores with \textit{b} if $R$ and $E$ were of the same size (balanced) and \textit{imb} in the opposite case (imbalanced). In all experiments, the components analyzed in the local evaluation were sorted by their size in decreasing order such that $\mathcal{G}_0$ always denotes the largest component in the graph.

\section{Experiment 1: Contrastive Learning} 
 \label{sec:exp:multiview-box}
\begin{wrapfigure}{l}{0.5\columnwidth}
 \begin{center}
    {\includegraphics[width=0.5\columnwidth]{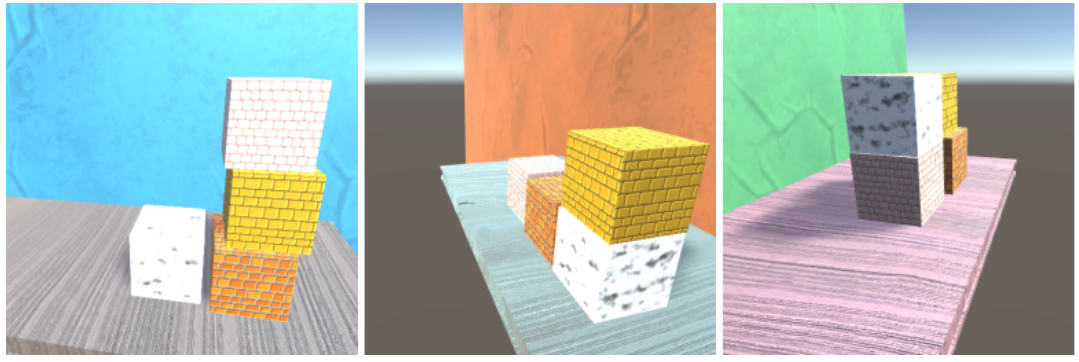}
    \caption{Examples of box images recorded from front, right and left views (left to right) contained in $\mathcal{D}_f$ and $\mathcal{D}_m$. }
    \label{fig:boxes-examples}}
    \end{center}
\end{wrapfigure}

We evaluated two models for learning contrastive representations, Siamese and SimCLR, on an image dataset %representing a box stacking task
introduced by \cite{state_repr_robotics}. Images, shown in Figure \ref{fig:boxes-examples}, 
consist of four boxes placed in $12$ possible arrangements recorded from front, left and right camera views in different scene color configurations. In this experiment, we used two of their datasets: {(i)} $\mathcal{D}_f$ containing front view images %(Figure \ref{fig:boxes-examples}, left column), 
and {(ii)} $\mathcal{D}_m$ additionally containing images recorded from the left and right views. %(Figure \ref{fig:boxes-examples}, middle and right, respectively). 
Each dataset consists of $5000$ training images and $5000$ test images not used during training. 
%There is total $12$ possible arrangements of the boxes which determine the label $c$ of each image. We compared Siamese and SimCLR models trained on $\mathcal{D}_f$ and $\mathcal{D}_m$, each consisting of $5000$ training images of size $256\times256$. The authors additionally provided $5000$ test images not used during training.  In this experiment, 
We always constructed $R$ and $E$ from $12$-dimensional representations of training and test images, respectively.

\begin{figure}[t]
\vskip 0.2in
\centering
\begin{center}
\includegraphics[width=1\columnwidth]{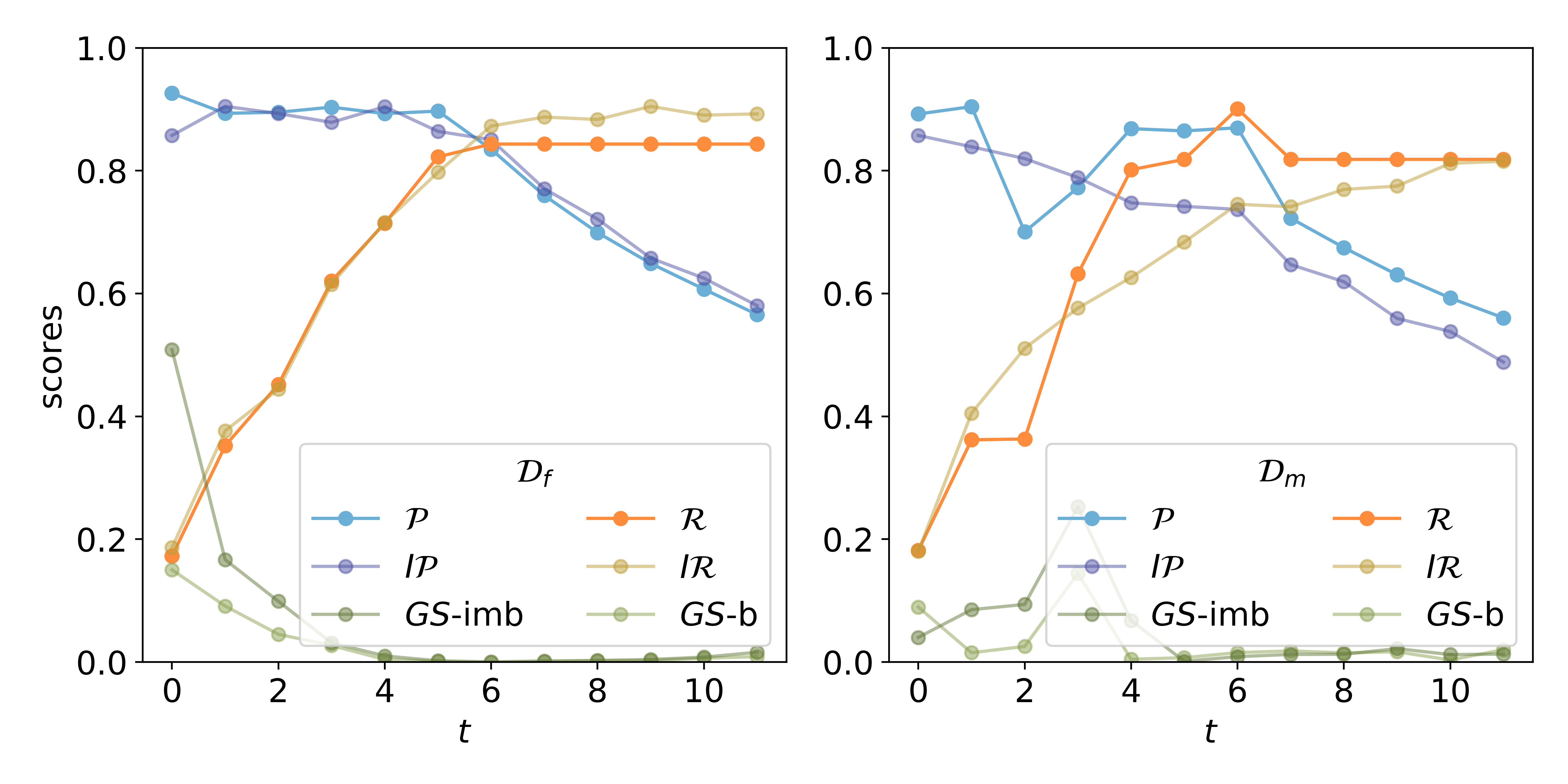} % first figure itself
\caption{Precision $\mathcal{P}$ and recall $\mathcal{R}$ scores obtained on representations from Siamese network trained on $\mathcal{D}_f$ (left) and $\mathcal{D}_m$ (right) when varying mode truncation level $t$. We compare the results with
$I\mathcal{P}$ and $I\mathcal{R}$ scores, as well as GS computed on balanced (\emph{b}) and imbalanced (\emph{imb}) sets $R$ and $E_t$ (multiplied by 10 on the right).}
\label{fig:box-truncation}
\end{center}
\vskip -0.2in
\end{figure}

\textbf{Mode Truncation Experiment}
In the first experiment, we applied GeomCA to investigate mode collapses and mode discoveries, two possible scenarios occurring during training of deep neural networks. We constructed $R$ from representations corresponding to the first $7$ classes (arrangements of boxes), $c_0, \dots, c_6$, %from each considered dataset, $\mathcal{D}_f$ and $\mathcal{D}_m$. We then 
and defined the sets $E_t$ to contain images from the first $t$ classes,  $c_0, \dots, c_t$, for $t = 0, \dots, 11$ (see %Supplementary Material 
Appendix~\ref{app:box} 
for exact sizes of these sets). 
Therefore, $E_t$ imitate mode collapse for $t < 7$ and mode discovery for $t > 7$. Since contrastive learning models should
%produce dense clusters corresponding to representations of the same class,
encode similar classes closeby, 
we used a small $\varepsilon = \varepsilon(1)$. Moreover, we used $\delta = \frac{\varepsilon}{2}$ to allow the homogeneous clusters also forming a component (see discussion in Sections \ref{sec:epsilon_selection} and~\ref{sec:gamma}), and chose $\eta_c = 0.75, \eta_q = 0.45$ in order to analyze only consistent components of high quality.

In Figure \ref{fig:box-truncation}, we show precision and recall scores, $\mathcal{P}, \mathcal{R}$, obtained on $R \cup E_t$ for each $t$. %using $\eta_c = 0.75, \eta_q = 0.45$. 
The representations were obtained from two Siamese models trained on $\mathcal{D}_f$ (left) and $\mathcal{D}_m$ (right). We observe that the scores correctly reflect the number of modes covered by each $E_t$, where recall (in orange) increases but precision (in blue) decreases with increasing $t$. At $t = 6$, where $E_6$ perfectly matches $R$, both $\mathcal{P}, \mathcal{R}$ are high.
We observe that the scores correlate well with the IPR scores (visualised in purple and yellow), but not with GS which fails to detect mode discovery cases. Note that IPR scores require $R$ and $E$ to have the same size and were obtained by randomly sampling $\min(|R|, |E|)$ samples for each of the sets. This is not needed for GeomCA which can handle even heavily imbalanced sets, for example, as obtained for $t = 0$ or $t = 11$. We applied GS on both imbalanced (dark green) and balanced (light green) sets, following the authors' recommendation, which yielded similar results.  Moreover, GeomCA and IPR correctly identify the modes even on the harder dataset $\mathcal{D}_m$ (right panel), where GS is unsuccessful also for the mode collapse cases. %which is on par with the results presented in \cite{state_repr_robotics}. % with a larger variation due to the increased complexity of images.
%reflect the modes even on the harder dataset $\mathcal{D}_m$. 
%This is on par with the results presented in \cite{state_repr_robotics}, where Siamese network is shown to be able to separate the representations by their classes even for $\mathcal{D}_m$. 

\begin{figure}[t]
\vskip 0.2in
\centering
\begin{center}
\includegraphics[height=0.5\columnwidth]{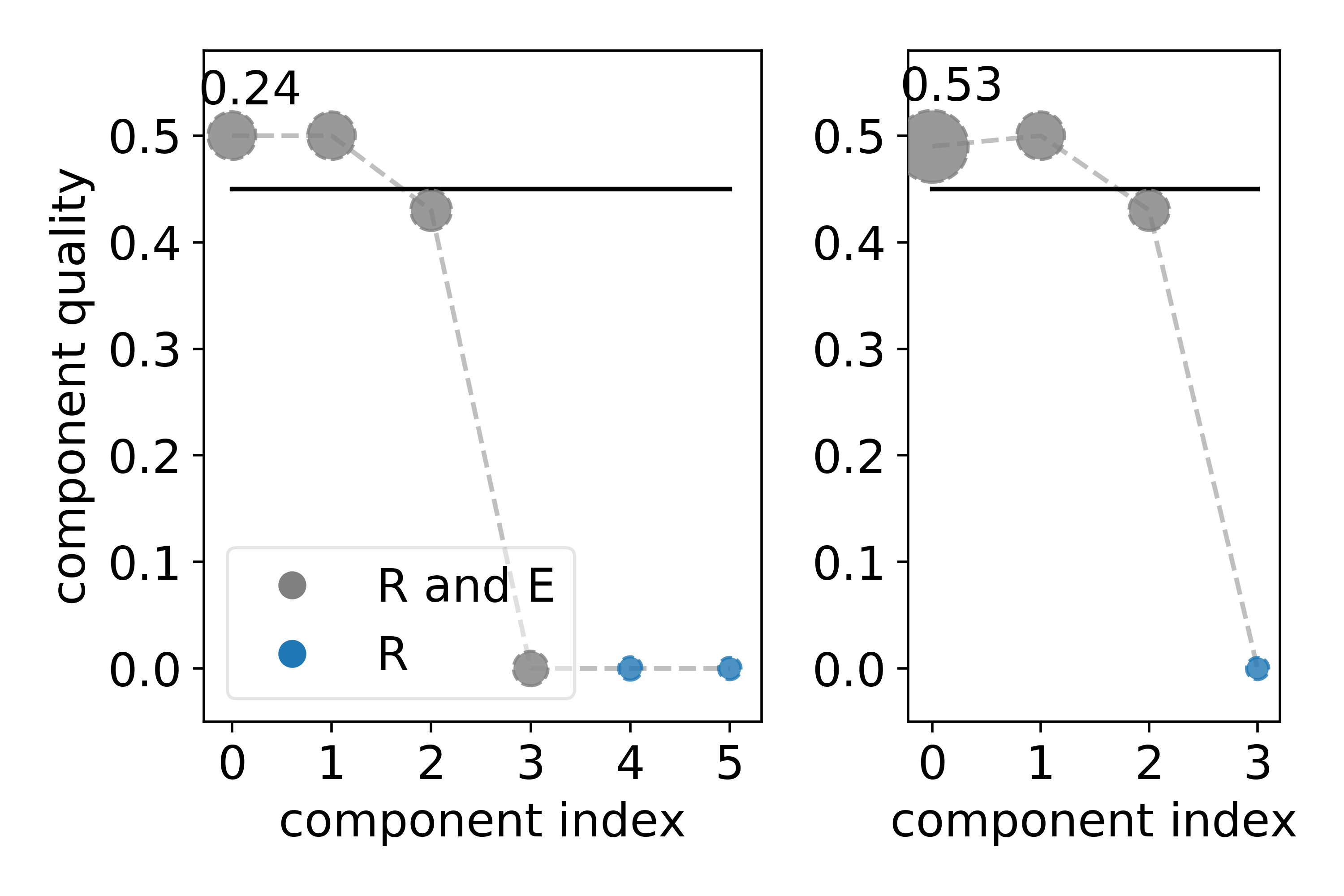} 
\caption{Quality of the components (y-axis) containing more than $100$ points obtained at $t = 2$ (left) and $t = 3$ (right) from the Siamese network on $\mathcal{D}_m$. The gray line denotes the threshold $\eta_q = 0.45$. Components' markers are scaled with their size. Gray denotes heterogeneous components, while blue denotes homogeneous components consisting only of $R$.}
\label{fig:box-class34}
\end{center}
\vskip -0.2in
\end{figure}

However, using local evaluation, GeomCA can provide a more detailed insight into the sets $R$ and $E_t$. For example, in the right panel of Figure \ref{fig:box-truncation} we observe a drop in $\mathcal{P}, \mathcal{R}$ scores at $t = 2$. % for the model trained on the harder dataset $\mathcal{D}_m$. 
We investigated this %drop
by analyzing the quality of the components $\mathcal{G}_i$ of $\mathcal{G}(R \cup E_2)$ containing at least $100$ representations, i.e, $|\mathcal{G}_i|_\mathcal{V} > 100$. The resulting scores are visualized in the left panel of Figure \ref{fig:box-class34}, where the markers of the components were scaled with their size, and colored with blue if they contain only points from $R$ and gray if they additionally contain points from $E$.
% We visualizethe obtained quality scores for each such component of the graph $\mathcal{G}(R \cup E_2)$ obtained at $t = 2$ in the left panel of Figure \ref{fig:box-class34}. The marker of each of the $5$ components is scaled with the size of the component, and colored with blue if the component contains only points from $R$ and gray if it additionally contains points from $E$. 
We clearly see that the drop in $\mathcal{P}, \mathcal{R}$ scores originates from the large heterogeneous component $\mathcal{G}_2$ with $q(\mathcal{G}_2)$ just below the chosen threshold $ \eta_q = 0.45$. Moreover, we see that the model fails to fully separate all $7$ classes since we can observe only $6$ large components (x-axis). This is even more evident when performing the same analysis on $\mathcal{G}(R \cup E_3)$ visualized in the right panel of Figure \ref{fig:box-class34}. Here, we observe only $4$ large components and a significant growth in size of the first component which now contains 53\% of $R \cup E_3$ instead of 24 \% of $R \cup E_2$ as for $t = 2$. 
% ADDED
Note that such detailed insights cannot be obtained by the existing frameworks such as IPR and GS.

%This experiments demonstrates how GeomCA can provide useful insights into the representations space that cannot be extracted only from $\mathcal{P}, \mathcal{R}$ scores.

In %Supplementary Material, 
Appendix \ref{app:box}, 
we demonstrate how the component consistency and quality thresholds $\eta_c, \eta_q$ can be flexibly adjusted to evaluate only components of certain minimum quality.

\textbf{Evaluating class separability}
In Figure~\ref{fig:box-truncation} (left), we have seen that $R$ and $E_6$ obtained from $\mathcal{D}_f$ by the Siamese model are well aligned. In this experiment, we applied GeomCA to both Siamese and SimCLR models and investigated the extend of the separation that these models achieve among the $7$ classes contained in the sets $R$ and $E_6$. 
In an ideal case, we would observe exactly $7$ \textit{consistent} components of \textit{high-quality}. Moreover, if the clusters are far apart from each other, the result should be robust to the choice of the distance threshold $\varepsilon$. % the components would be visible for a large range of $\varepsilon$ choices. 
However, as discussed in Section \ref{sec:epsilon_selection}, for a too small $\varepsilon$, there should be no such components, while for a too large $\varepsilon$ we should observe only one large component.
To eliminate the effect of sparsification and ensure perfect consistency, we randomly sampled $250$ points from each of the classes $c_t$ and ran GeomCA without any further reduction of points.

%\begin{wrapfigure}{l}{0.5\columnwidth}
% \begin{center}
%{\includegraphics[width=0.5\columnwidth]{Figures/box_epsilon_separability_onlysize.png}
    %\caption{Number of components containing more than $100$ points (y-axis) obtained from Siamese and SimCLR models when varying the distance threshold $\varepsilon$. }
    %\label{fig:boxes-separability}}
    %\end{center}
%\end{wrapfigure} 

\begin{figure}[h]
\vskip 0.2in
\centering
\begin{center}
\includegraphics[height=0.5\columnwidth]{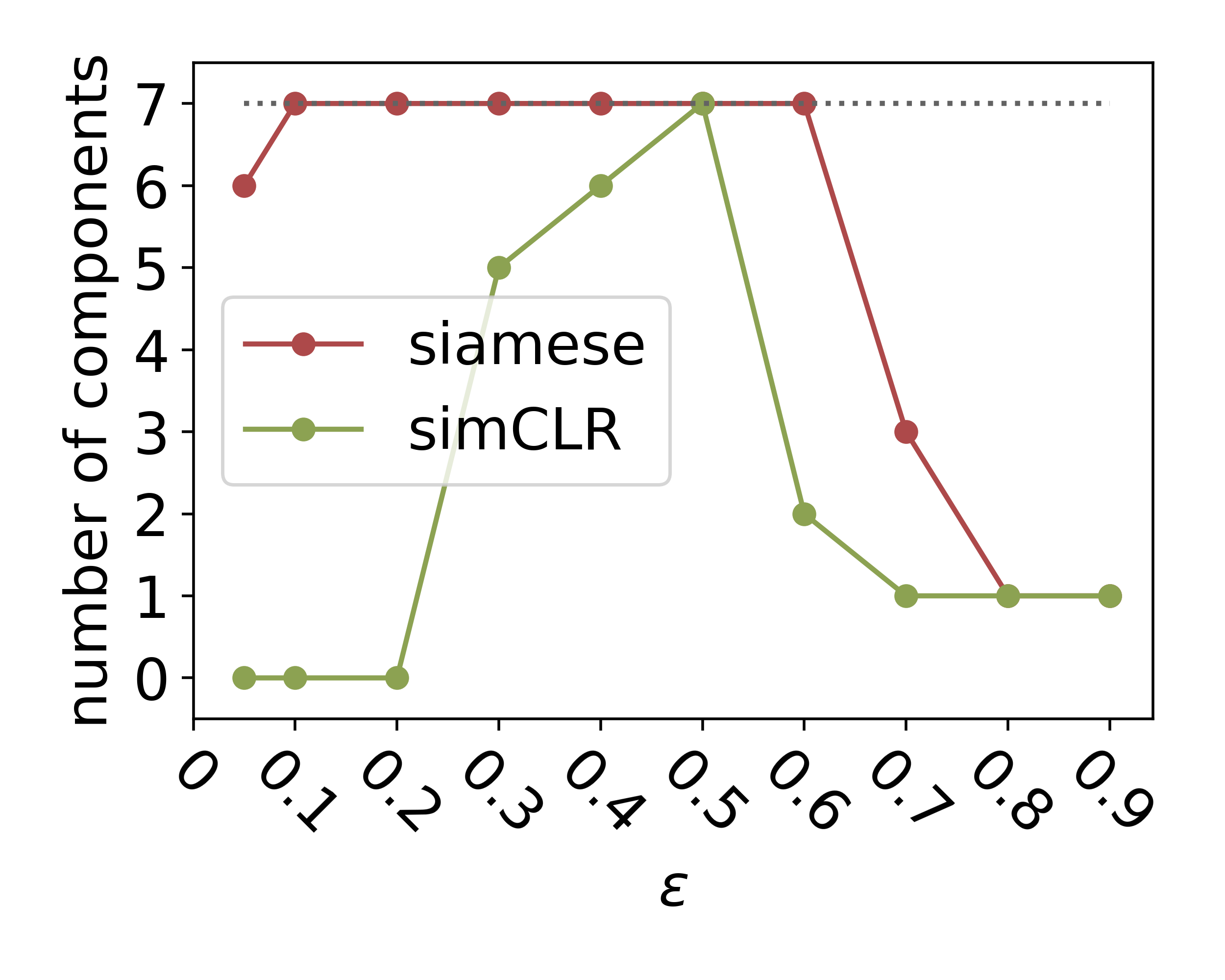} 
\caption{Number of components containing more than $100$ points (y-axis) obtained from Siamese and SimCLR models when varying the distance threshold $\varepsilon$. \\}
\label{fig:boxes-separability}
\end{center}
\vskip -0.2in
\end{figure}

In Figure \ref{fig:boxes-separability}, we plot the number of components with %$c(\mathcal{G}_i) > 0.75, q(\mathcal{G}_i) > 0.45, 
$|\mathcal{G}_i|_{\mathcal{V}} > 100$ obtained by Siamese (in green) and SimCLR models (in red) when varying $\varepsilon \in \{ 0.05, 0.1, \dots, 0.9\}$. We clearly observe that only Siamese model well separates the classes since we observe $7$ components for a large range of $\varepsilon$ choices.
Surprisingly, SimCLR is much more sensitive to the choice of $\varepsilon$, Moreover, we observe that Siamese network achieves higher network quality than SimCLR 
(visualized in Figure~\ref{fig:box-separability-network_quality}, Appendix~\ref{app:box}),

\section{Experiment 2: Generative Models} \label{sec:exp:gan}
GeomCA algorithm can also be used to evaluate the quality and diversity of samples generated by generative models. We used a StyleGAN trained on FFHQ dataset \cite{stylegan} and replicated the truncation experiment as performed in \cite{ipr}. Here, the latent vectors generating images are during testing sampled from a truncated normal distribution such that the values which fall outside a given range are resampled to fall inside that range~\cite{truncation}. The level of truncation is controlled by the parameter $\psi$ determining a tradeoff between perceptual quality and variation of images.
%Here, the latent vectors generating images are truncated with a given parameter $\psi$ determining a tradeoff between perceptual quality and variation of images. %Using the official code provided by \cite{ipr}, 
We generated $50000$ images and obtained their $4096$-dimensional representations from a pretrained VGG16 model. These composed the set $E$, while we created $R$ from $50000$ representations of the training data. Due to large dimensionality of the representations, we chose $\varepsilon = \varepsilon(10)$, and $\eta_c, \eta_q = 0$. Since the generated representations $E$ are in an ideal case well aligned with $R$, we chose $\delta = \varepsilon$.

%We generated $50000$ images %from the pretrained StyleGAN  and obtained their $4096$-dimensional representations from a pretrained VGG16 model, using the official code provided by \cite{ipr}. We defined $E$ to contain the generated representations, while we created $R$ from $50000$ representations of the training data. We then ran GeomCA and compared our $\mathcal{P}, \mathcal{R}$ scores with IPR and GS. Due to large dimensionality of the representations, we chose $\varepsilon = \varepsilon(10)$, and $\eta_c, \eta_q = 0$. Since we wish the generated representations $E$ to be well aligned with $R$, we chose $\delta = \varepsilon$.

\begin{figure}[ht]
\vskip 0.2in
\begin{center}
\centerline{\includegraphics[width=\columnwidth]{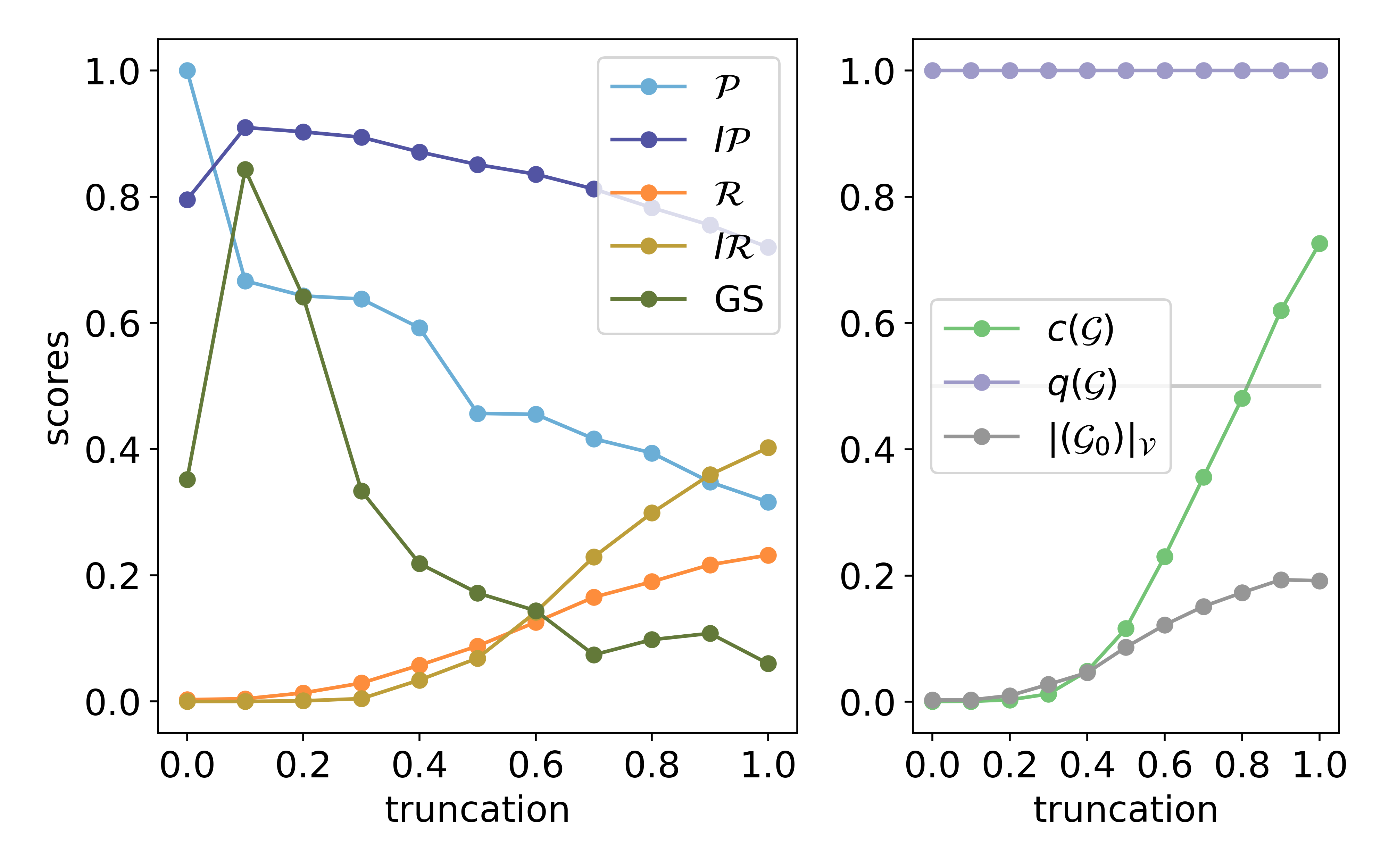}}
\caption{Results of the StyleGAN truncation experiment. Left: GeomCA precision and recall $\mathcal{P}, \mathcal{R}$ compared with IPR and GS (multiplied by 10) scores. Right: network consistency $c(\mathcal{G})$ and quality $q(\mathcal{G})$ as well as the size $|\mathcal{G}_0|_\mathcal{V}$ of the only component containing more than $100$ points (scaled by the number of all points in $\mathcal{G}$).}
\label{fig:styleGAN-truncation50k}
\end{center}
\vskip -0.2in
\end{figure}

In the left panel of Figure \ref{fig:styleGAN-truncation50k}, we visualize GeomCA $\mathcal{P}$ and $ \mathcal{R}$, IPR and GS (multiplied by 10) scores obtained at each truncation level $\psi$. %We again observe GeomCA and IPR to reflect the applied truncation, while this is not the case with GS. 
We observe that all methods reflect the applied truncation, with some deviations for GS at $\psi = 0$. Comparing GeomCA and IPR, we observe fairly consistent recall but more variation in the precision scores. Therefore, %To gain more insight into the structure of representations, 
we further investigated the network consistency $c(\mathcal{G})$ and network quality $q(\mathcal{G})$. The results, visualized in the right panel of Figure \ref{fig:styleGAN-truncation50k}, show that the network consistency (green) is lower than $0.5$ for $\psi \le 0.8$. This is the effect of the geometric sparsification, which in fact removes the majority of $E$ points. For example, the sparsified $E$ contains only $2$ points for $\psi = 0.0, 0.1$. As in Section \ref{sec:gamma}, we argue that this provides valuable insights into the structure of the generated points. If these reflected the structure of the training points $R$, the sparsification would return sparsified sets $R$ and $E$ of approximately the same size. Since this is not the case even for $\psi = 1.0$, we argue that the model fails to fully learn the true distribution of the training data, which is also reflected in our low precision scores $\mathcal{P}$.
Note that the network has perfect quality regardless the value of $\psi$ due to $\delta = \varepsilon$. As discussed in Section \ref{sec:gamma}, this requires every point in $R$ to be `witnessed'' by a point in $E$, which give rise to heterogeneous edges in the network.

%Secondly, we observe that the network has perfect quality regardless the value of $\psi$. This means that the only edges in the graph are the heterogeneous ones connecting $R$ and $E$ points. We argue that this is beneficial since it means that $E$ is well mixed with $R$. In other words, for every point in $R$ there is a $E$ point "witnessing" it as discussed in Section \ref{sec:gamma}. 

Investigation of homogeneous edges requires to choose $\delta < \varepsilon$ and potentially increasing $\varepsilon$ itself.
%The constant network quality on the other hand means that there are no edges among $R$ points, which could happen for a small $\varepsilon$ or large sparsification parameter $\delta$. 
In %Supplementary Material, 
Appendix \ref{app:gen_models}, 
we provide further experiments when varying both $\varepsilon$ and $\delta$ and show that GeomCA correctly reflects the structure of the space in all cases. We also use this large scale experiment to perform both time complexity and robustness analysis for varying number of samples considered in the sets $R$ and $E$. %These results can be found in Appendix CITE.

\section{Experiment 3: VGG16 Model} \label{sec:exp:vgg16}
The FFHQ representations in the StyleGAN evaluation in Section \ref{sec:exp:gan} are obtained from a VGG16 model pretrained on the ImageNet dataset. In a detailed analysis, we always observed only one connected component containing more than $100$ points, the size of which grew with the truncation $\psi$ as shown in the right panel of Figure \ref{fig:styleGAN-truncation50k} (in gray and labeled with $|\mathcal{G}_0|_\mathcal{V}$).
%, we can additionally see that the size of the largest component $|\mathcal{G}_0|_\mathcal{V}$ increases with the truncation $\psi$. We observe that the FFHQ representations in fact always composed one large connected component.
However, since VGG16 is a supervised learning model, we would expect it to be able to separate representations at least to some extent. %However, we observed that the FFHQ representations composed one large connected component, the size of which increased with increasing truncation $\psi$ [CITE APPENDIX AND VISUALIZE]. 
%It is possible that the features of face images contained in FFHQ dataset are not as distiguishable as in other datasets, such as ImageNet or $\mathcal{D}_f, \mathcal{D}_m$ from Section \ref{sec:exp:multiview-box}. 
To determine whether this inseparability originates from the VGG16 model or the nature of the FFHQ dataset, which contains images of faces, we also applied GeomCA to VGG16 representations of the ImageNet dataset.

We performed a simple experiment and defined the sets $R$ and $E$ to contain $5$ different classes of the ImageNet dataset each. In version $1$, we manually chose  classes representing kitchen utilities for $R$, and dogs for $E$ such that $R$ and $E$ contain semantically different representation (see Appendix \ref{app:vgg16_model} 
%Supplementary Material 
for exact labels and sizes). In version $2$, the $5$ classes for $R$ and $E$ were chosen at random. If VGG16 is able to separate the classes, then we do not expect to obtain components with high consistency and quality in version $1$, while few small ones can emerge in version $2$ due to the random choice. Moreover, if the sets in version $1$ reflect differences in semantic information of classes, this could potentially be seen in the imbalances after the sparsification process, while this should not be significant in version $2$. 
%In the second version, we can expect few small components with high consistency and quality. In contrast to version one, the sparsification should result in approximately the same sizes of $R$ and $E$, exactly because of the random choice.

As in Section \ref{sec:exp:gan}, we estimated $\varepsilon = \varepsilon(10)$ and chose $\delta = \varepsilon$. The results of the global and local GeomCA evaluation as well as IPR and GS scores are shown in Table \ref{tab:vgg19}. The graph $\mathcal{G}$ in version $1$ achieves $75 \%$ consistency, which is indeed the result of sparsification. This indicates that the designed sets $R$ and $E$ contain different semantic information. Moreover, since $\mathcal{P}$ and $\mathcal{R}$ are both low, we conclude that there is little overlap between the sets. This can also be seen from the fact that we obtain only $7$ non trivial components containing more than one point, where the largest component $\mathcal{G}_0$ contains only $18$ elements. 
On contrary, we observe high $c(\mathcal{G})$ and slightly larger $\mathcal{P}, \mathcal{R}$ scores in version $2$, which indicates that there are few areas where $R$ and $E$ are well aligned.
%have more similar topology but not geometry as seen from low $\mathcal{P}$, $\mathcal{R}$ scores. 
%In contrast to the first version, we also observe 
This can also be seen from larger number of non trivial components as well as more points in the largest component. %This suggests that $R$ and $E$ are better aligned as in version $1$ but still quite separated.
Note that $q(\mathcal{G}) = 1$ due to $\delta = \varepsilon$. 
Therefore, we hypothesise that VGG16 achieves a certain level of separation of ImageNet training classes. We emphasise that it is difficult to draw the same conclusion from either IPR or GS as they fail to provide such detailed insight.

\begin{table}[t]
\caption{GeomCA scores obtained on VGG16 representations from ImageNet experiment in version $1$ (Kitchen utilities vs dogs) and version $2$ (random) compared with IPR and GS scores.}
\label{tab:vgg19}
\vskip 0.15in
\begin{center}
\begin{small}
\begin{sc}
\begin{tabular}{rcc}
\toprule
 & kitchen vs. dogs & random \\
\midrule
$c(\mathcal{G}), q(\mathcal{G})$    &  \num[round-mode=places,round-precision=2]{0.745747735807378}, \num[round-mode=places,round-precision=2]{1.0}  &
\num[round-mode=places,round-precision=2]{0.9825195901145268}, \num[round-mode=places,round-precision=2]{1.0}
\\
%$q(\mathcal{G})$   &  \num[round-mode=places,round-precision=4]{1.0} & \num[round-mode=places,round-precision=4]{1.0} \\
$\mathcal{P}$   & \num[round-mode=places,round-precision=4]{0.004226840436773511}
& 
\num[round-mode=places,round-precision=4]{0.04233128834355828}\\
$\mathcal{R}$ & \num[round-mode=places,round-precision=4]{0.013033175355450236} 
& \num[round-mode=places,round-precision=4]{0.03909952606635071} 

\\
$|\mathcal{G}^R|_\mathcal{V}$ & 1688 & 1688 \\
$|\mathcal{G}^E|_\mathcal{V}$ & 2839 & 1630 \\

$|\mathcal{G}_0|_\mathcal{V}$ & 18 & 77 \\
%\makecell{
$\#$ non trivial $\mathcal{G}_i$ & 7 & 25 \\
\midrule
$I\mathcal{P}, I\mathcal{R}$ & 
\num[round-mode=places,round-precision=2]{0.7780441}, \num[round-mode=places,round-precision=2]{0.96963886}  
& 
\num[round-mode=places,round-precision=2]{0.94630769} , \num[round-mode=places,round-precision=2]{0.98369231}  \\
GS & \num[round-mode=places,round-precision=4]{0.0018294223910747193}  & \num[round-mode=places,round-precision=4]{0.00041750875851729707} \\

\bottomrule
\end{tabular}
\end{sc}
\end{small}
\end{center}
\vskip -0.1in
\end{table}

% T: chambered nautilus, shopping basket, leafhopper, ski,altar
% V: rock crab, slide rule, ice bea, basset, ladle

\begin{figure}[t]
\vskip 0.2in
\begin{center}
\centerline{\includegraphics[width=0.95\columnwidth]{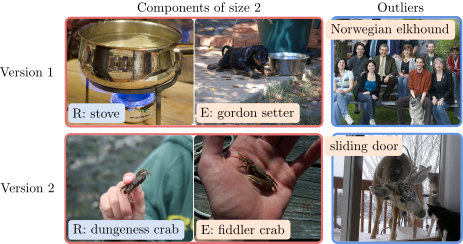}}
\caption{Examples of ImageNet images corresponding to the representations from version $1$ (top) and version $2$ (bottom) obtained from a pretrained VGG16. Images with red stroke are taken from components of size $2$, where left column images belong to representations of $R$ and right ones to $E$. 
Images with blue stroke correspond to outliers. % from version $1$ (top) and version $2$ (bottom).
%In the right column, we show two outliers from the sets $E$. Top one corresponds to version $1$, bottom one to version $2$. 
See the text for discussion. 
}
\label{fig:method:imagenet_examples}
\end{center}
\vskip -0.2in
\end{figure}

To gain further insights into separability capabilities of the model, we visualize images of representations contained in the obtained components. In Figure \ref{fig:method:imagenet_examples}, we visualize in red a component containing two representations, and in blue an $E$ outlier, both from version $1$ (top row) and version $2$ (bottom row). In both cases, red components show examples of erroneous merges based on human labels, which are %on the other hand 
not surprising due to the striking similarity between the images. In the top row, both images contain a silver pot, while the right one also shows a dog.
%Lastly, we demonstrate how GeomCA can be used to investigate individual components. 
%On the left, we visualize two components (red) each containing $2$ representations, obtained in version $1$ (top) and $2$ (bottom). The left images belong to the set $R$, while the right ones belong to the set $E$. Both rows show examples of erroneous merges, which are on the other hand not surprising due to the striking similarity between the pairwise images. In top row, both images contain a silver pot, while the right one also shows a dog. 
In the bottom row, the crabs in the images belong to two different species that are arguably hard to differentiate but both are placed in a human hand. Another example from version $2$ is shown in Figure \ref{fig:exp:vgg16-front} where images have similar background but contain different object in the center.  %Lastly, in panel (c), we visualize images corresponding to isolated $E$ representations from version $1$ (top) and version $2$ (bottom). 
In the case of outliers, it is rather hard to spot the dog in the lower corner of the top row image, while the sliding window in the bottom one, which is the image label, seems to be of secondary focus after the animals.

\section{Conclusion and discussion}
We presented GeomCA algorithm for evaluating topological and geometrical properties of representation spaces. The intuition behind GeomCA is that if two given sets of representations $R$ and $E$ contain observations from the same data manifold, then they are necessarily well aligned. We measure this alignment by analyzing the connected component of an $\varepsilon$-threshold graph built on their union $R \cup E$. For each component, we determine its \textit{consistency} by measuring the ratio of points from $R$ and $E$ contained in it, and \textit{quality} by measuring the ratio of heterogeneous edges connecting points from $R$ and $E$. Moreover, we aggregate these scores into four global measures, \textit{precision, recall, network consistency} and \textit{network quality}. We demonstrate the usefulness of the proposed global and local measures in several different scenarios such as  evaluation of separability of representations obtained from both contrastive learning or supervised learning algorithms as well as in the evaluation of trained generative models.

%[DISCUSS LIMITATIONS?] Should we discuss limitations of the algorithm, such as components with different densities. Another thing is the severity of sparsification: if it is random [CHECK], then it might remove outliers that we want to be aware of.

\newpage 
%\clearpage
\bibliography{references}
\bibliographystyle{icml2021}

% Uncomment this to get camera ready          
%\newpage
%\clearpage
\appendix
\section{Experimental details} \label{app:experiments}
In this section, we provide information about the hyperparameters used for GS and IPR methods as well as further experimental results supporting the conclusions in the main part.

We always evaluated IPR using neighborhood size $k = 3$ as suggested by the authors. For this, we used balanced sets $R$ and $E$ obtained by sampling $\min(|R|, |E)$ points from each of them. The hyperparameters used for GS are adjusted to the specific experiment and discussed in the sections below.

\subsection{Contrastive Learning} \label{app:box}

\textbf{Mode truncation experiment} The number of representations corresponding to each class $c_t$ for $t = 0, \dots, 11$ in the training and holdout splits of both $\mathcal{D}_f$ and $\mathcal{D}_m$ are shown in Table~\ref{app:tab:box-classes} (middle rows). The set $R$ was composed of representations corresponding to the first $7$ classes $c_0, \dots, c_6$ from the training split, which amounts to $3514$ points. The respective sizes of the sets $E_t$ are shown in the right column of the table. %Note that the sizes of $R$ and $E$ after sparsification slighly vary. 
The $\varepsilon(1)$ threshold evaluated to $0.05$ and $0.18$ in case of $\mathcal{D}_f$ and $\mathcal{D}_x$, respectively. Note that the value is constant across all values of $t$ because it was always estimated on $R$ using same random seed. In this experiment, we evaluated GS using $L_0 = 64$, $\gamma = 1/128$, $i_{\max} = 10$ and $n = 1000$.

\begin{table}[h]
\caption{Number of representations corresponding to each class contained in the training and holdout splits of both $\mathcal{D}_f$ and $\mathcal{D}_m$ datasets (middle columns). The respective sizes of each set $E_t$ used in our experiments is shown in the right column.}
\label{app:tab:box-classes}
\vskip 0.15in
\begin{center}
\begin{small}
\begin{sc}
\begin{tabular}{r|cc|c}
\toprule
 class & train & holdout & $E_t$ \\
 \midrule
 $0$ & 670 & 666 & 666
         \\
 $1$ & 690 & 625 & 1291  \\
 $2$ & 395 & 373 & 1664  \\
 $3$ & 706 & 684 & 2348  \\
 $4$ & 349 & 429 & 2777  \\
 $5$ & 409 & 377 & 3154  \\
 $6$ & 295 & 309 & 3463  \\
 $7$ & 296 & 312 & 3775  \\
 $8$ & 292 & 310 & 4085  \\
 $9$ & 311 & 293 & 4378  \\
 $10$ & 258 & 279 & 4657  \\
 $11$ & 331 & 345 & 5002  \\
\bottomrule
\end{tabular}
\end{sc}
\end{small}
\end{center}
\vskip -0.1in
\end{table}

%\textbf{Mode truncation experiment with varying consistency and quality thresholds}
    
\begin{figure*}[ht]
    \begin{minipage}[b]{.25\textwidth}  
      \centering  
      \includegraphics[width=1\linewidth, height=0.2\textheight]{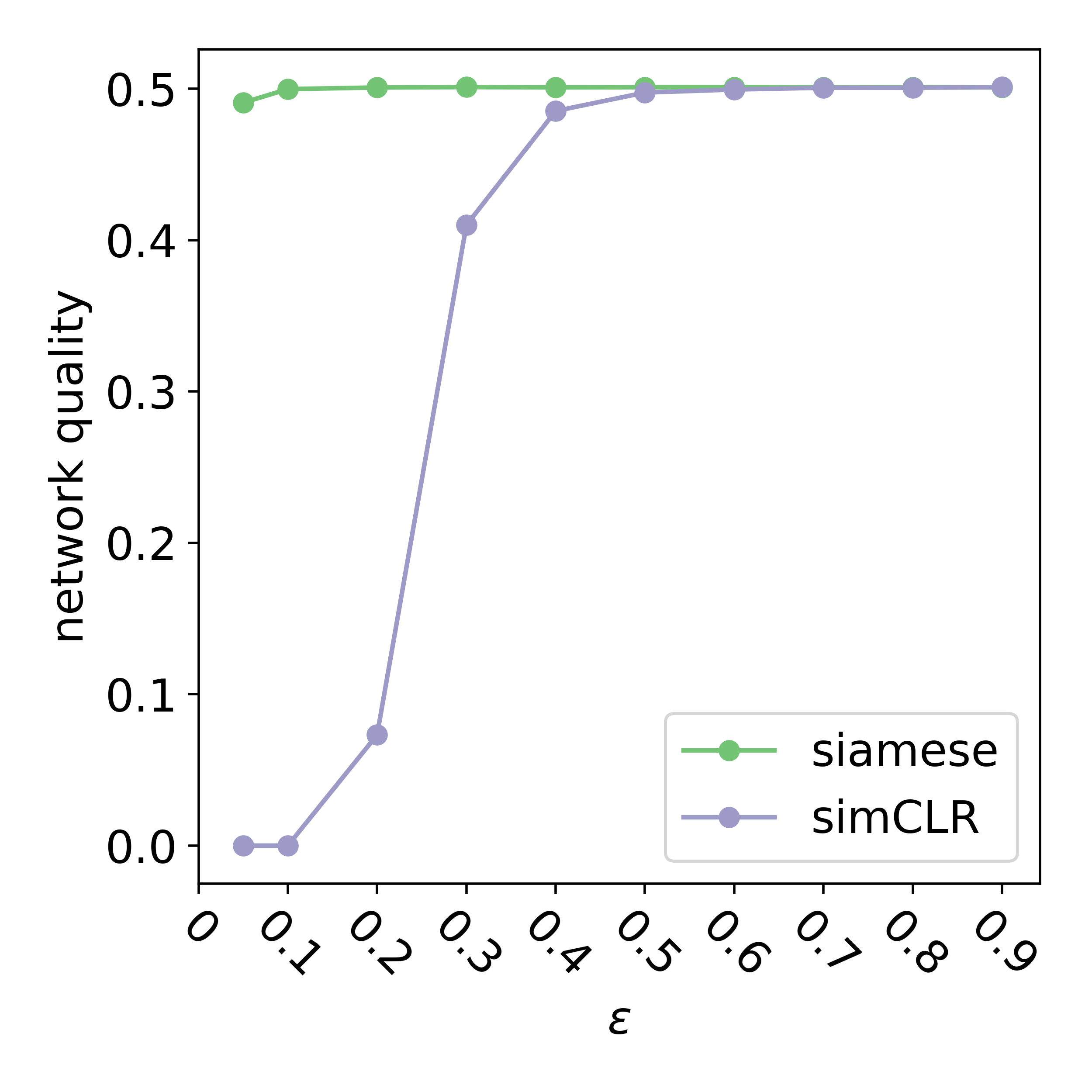}
    \end{minipage}%  
    \hfill
    \begin{minipage}[b]{0.65\textwidth}  
      \centering  
      \includegraphics[width=1\linewidth, height=0.2\textheight]{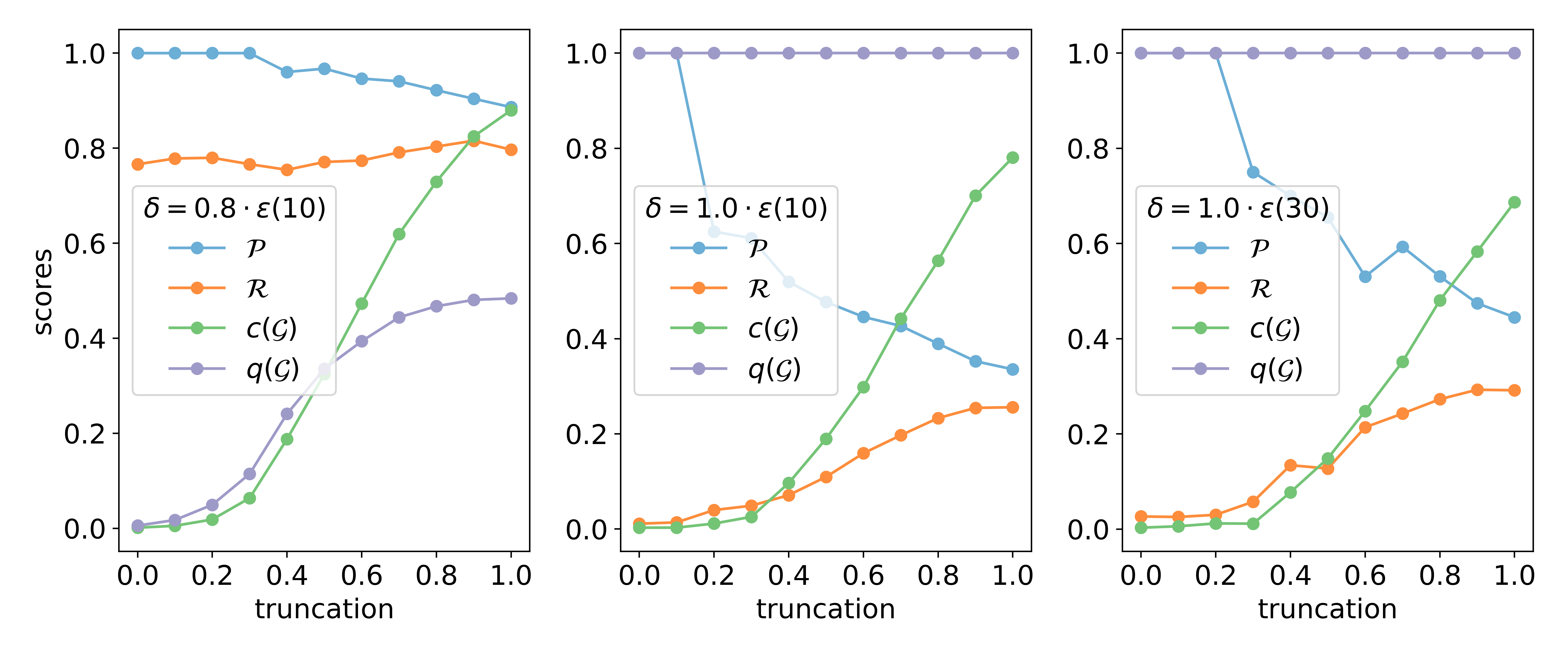}
    \end{minipage}
    \par
    \begin{minipage}[t]{.25\textwidth}
      \caption{Network quality $q(\mathcal{G})$ (y-axis) obtained for Siamese and SimCLR models on $\mathcal{D}_f$ when varying distance threshold $\varepsilon$. }
        \label{fig:box-separability-network_quality}  
      \label{bilayer}  
    \end{minipage}%
    \hfill
    \begin{minipage}[t]{.65\textwidth}  
      \caption{Precision $\mathcal{P}$, recall $\mathcal{R}$, network consistency $c(\mathcal{G})$ and network quality $q(\mathcal{G})$ obtained in the StyleGAN experiment when varying $\delta$ (left and middle) and when varying $\varepsilon$ (middle and right).}
        \label{fig:stylegan-eps-delta}
    \end{minipage}  
    \end{figure*}
    
\begin{figure}[ht]
\vskip 0.2in
\begin{center}
\centerline{\includegraphics[width=\columnwidth]{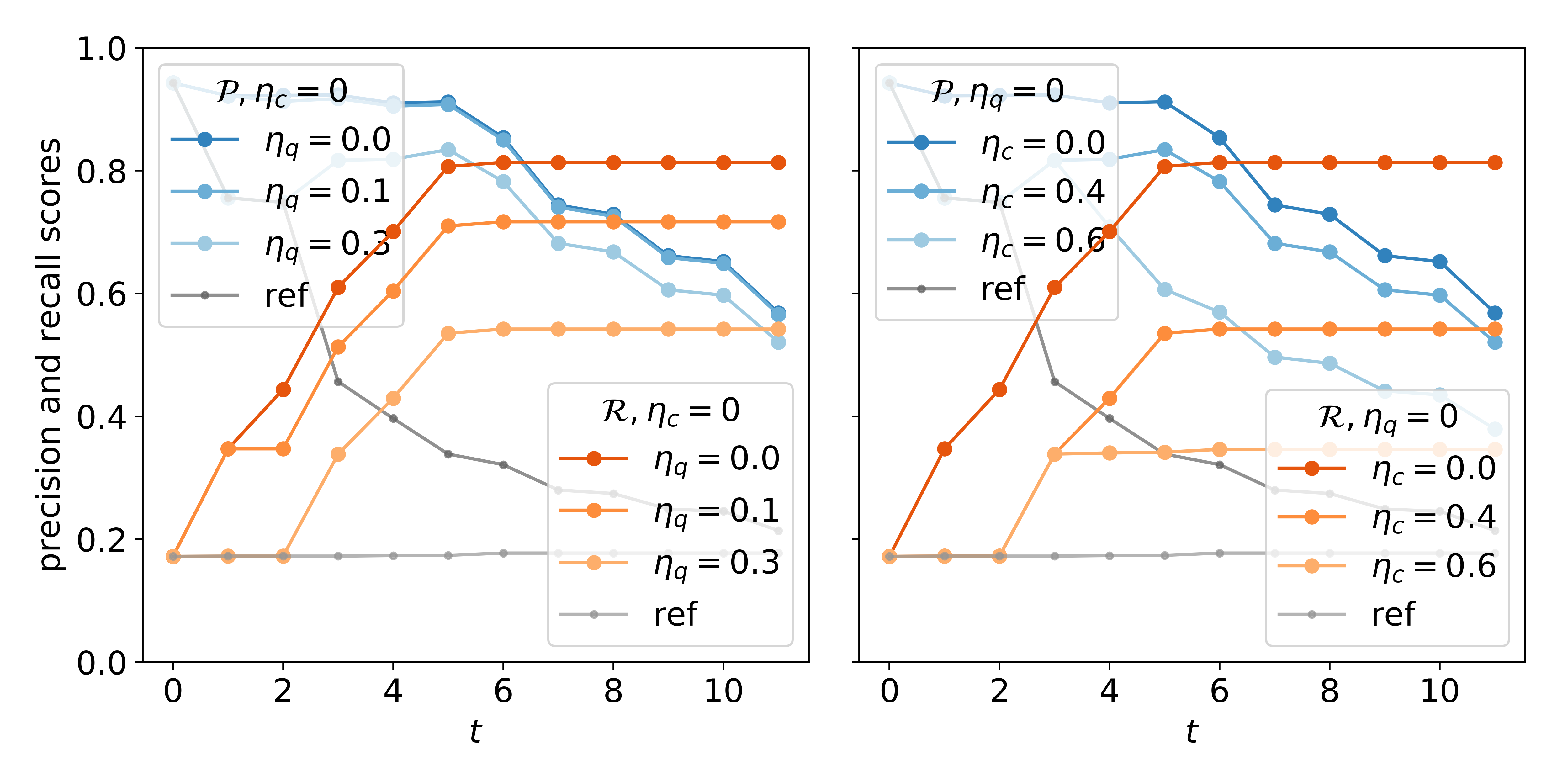}}
\caption{Precision and recall scores obtain when varying the consistency and quality thresholds $\eta_c, \eta_q$. Left: scores obtained for a fixed $\eta_c = 0$ and varying $\eta_q$. Right: scores obtained for a fixed $\eta_q = 0$ and varying $\eta_c$. Both panels show reference scores (gray) obtained using $\eta_c = 0.75$ and $\eta_q = 0.45$.}
\label{fig:box-ctcq}
\end{center}
\vskip -0.2in
\end{figure}

Next, we demonstrate how the variations in the component consistency and quality thresholds $\eta_c, \eta_q$, respectively, can be used to evaluate only components of certain minimum quality. Since the components in the mode truncation experiment have both high consistency and high quality, we deliberately corrupted the sets $E_t$ to obtain more inconsistent and homogeneous components. Instead of adding all images of the class $c_t$ to $E_{t-1}$, we sampled a subset of them of a randomly chosen size. The $\mathcal{P}, \mathcal{R}$ scores obtained by varying $t$ and $\eta_c, \eta_q$ are shown in Figure \ref{fig:box-ctcq}. In the left panel, we visualize the scores obtained at a constant $\eta_c = 0$ and $\eta_q \in \{ 0, 0.1, 0.3\}$. In the right panel, we instead fix $\eta_q = 0$ and vary $\eta_c \in \{0, 0.4, 0.6\}$. In both panels, we additionally plot reference $\mathcal{P}, \mathcal{R}$ scores (in gray) obtained for $\eta_c = 0.45$ and $\eta_q = 0.75$. We observe that the scores correctly decrease when considering larger threshold values. 

\textbf{Evaluating class separability}
In Figure~\ref{fig:box-separability-network_quality}, we show network quality $q(\mathcal{G})$ obtained when varying the distance threshold $\varepsilon$ on both Siamese and SimCLR models on $\mathcal{D}_f$. We observe that the Siamese network, in addition to having $7$ components containing more than $100$ points for $0.1 \leq \varepsilon \leq 0.6$ (Figure 7), %(Figure~\ref{fig:boxes-separability}),
also achieves higher network quality. This means that these components also contain many heterogeneous edges, indicating that $R$ and $E$ are also well geometrically positioned.

\subsection{Generative models} \label{app:gen_models}
In this experiment, $R$ always contained VGG16 representations of $50000$ training data points from the FFHQ dataset, while $E$ contained $50000$ representations corresponding to the images generated by a trained StyleGAN model. The threshold $\varepsilon(10)$ was estimated to $28.10$ for all $\psi$. As before, we evaluated IPR using neighbourhood size $k = 3$, while evaluated GS on $10000$ randomly sampled points using $L_0 = 64, \gamma = 1/1280, i_{\max} = 100$ and $n = 1000$ for GS following the authors' recommendations except for the value of $n$. Initially, we tried running GS with $n = 10000$ and using all $50000$ points but stopped the evaluation because of too slow computations (around 8 hours CPU time per truncation).

\begin{figure}[ht]
\vskip 0.2in
\begin{center}
\centerline{\includegraphics[width=\columnwidth]{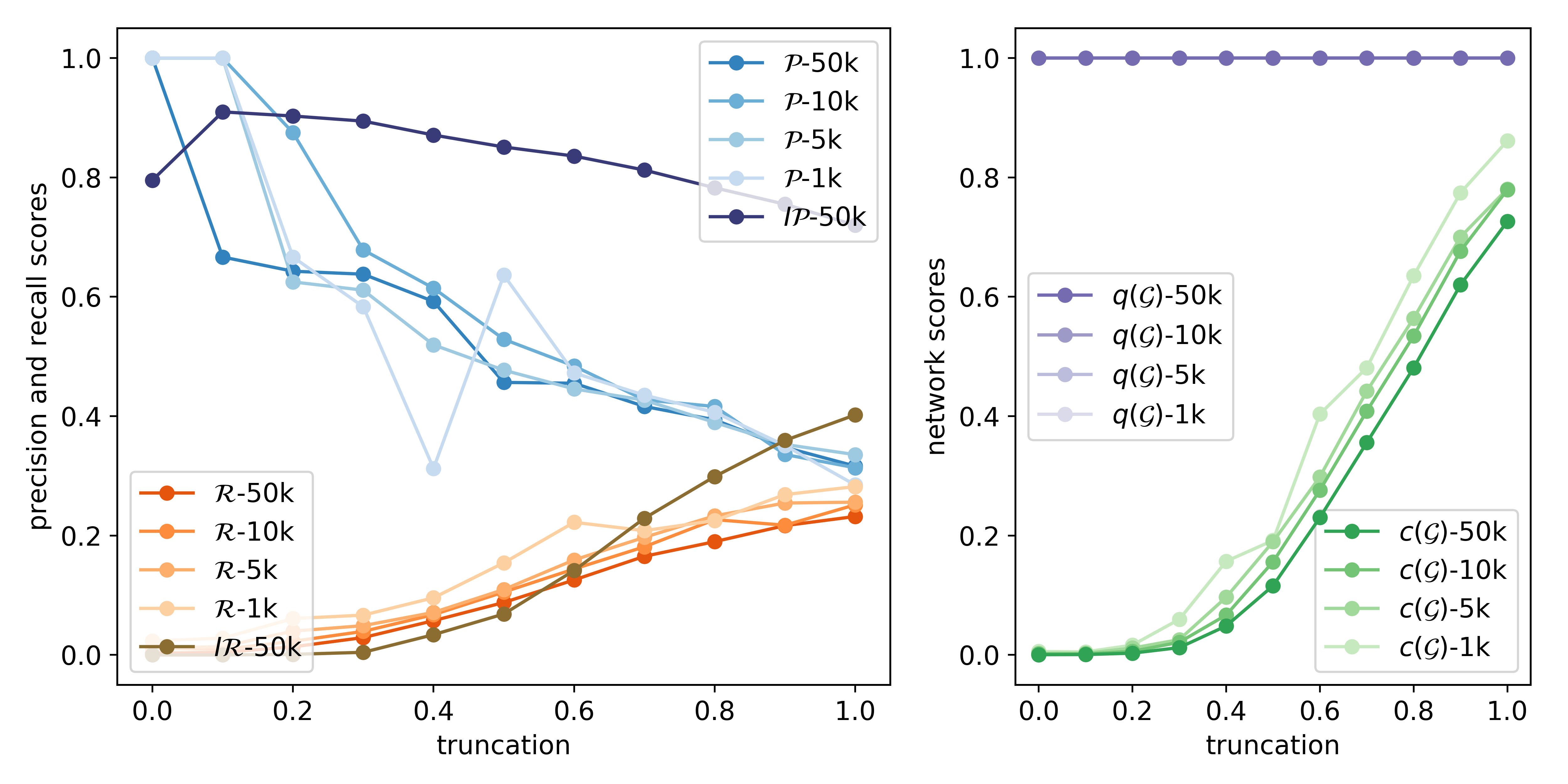}}
\caption{Results of StyleGAN truncation experiment obtained when varying the sizes of the sets $R$ and $E$ to contain $50000$ ($50k$), $10000$ ($10k$) $5000$ ($5k$), and $1000$ ($1k$) points. Left: GeomCA precision and recall scores $\mathcal{P}, \mathcal{R}$ as well as IPR scores obtained on $50000$ points. Right: GeomCA network consistency $c(\mathcal{G})$ and quality $q(\mathcal{G})$ scores.}
\label{app:fig:stylegan-samplessize}
\end{center}
\vskip -0.2in
\end{figure}

\textbf{Varying sample size}
We used this large-scale experiment to perform time complexity and robustness analysis for varying number of samples contained in the sets $R$ and $E$. For this, we additionally subsampled $10000, 5000$ and $1000$ samples from both $R$ and $E$ and calculated GeomCa global scores using $\delta = \varepsilon(10)$. In Figure~\ref{app:fig:stylegan-samplessize}, we plot the resulting $\mathcal{P}, \mathcal{R}$ scores (left panel) and $c(\mathcal{G}), q(\mathcal{G})$ scores (right panel) obtained on all sizes of $R$ and $E$. For comparison, we additionally visualize GeomCA and IPR obtained using all $50000$ points as in Section 5. %~\ref{sec:exp:gan}. 
We observe that GeomCA returns consistent results for all sizes except for the case of $1000$ points where we obtained slight inconsistencies for truncations $0.4 \leq \psi \leq 0.6$.

In Tables \ref{tab:exp:stylegan_sparsification_time} and \ref{tab:exp:stylegan_graph_time}, we report (CPU-based) time analysis of GeomCA obtained on truncation $\psi = 1.0$ corresponding to the varying sizes of the sets $R$ and $E$ as above. %In this experiment, we randomly subsampled $50000, 10000, 5000$ and $1000$ samples from each of the sets $R$ and $E$, and 
%We measured the sparsification time (Table \ref{tab:exp:stylegan_sparsification_time}) as well as time it took to build $\varepsilon$-graph on the resulting reduced sets $R'$ and $E'$ (Table \ref{tab:exp:stylegan_graph_time}). 
In Table \ref{tab:exp:stylegan_sparsification_time} we report the sizes (cardinality) of the $R$ and $E$ sets given as inputs to GeomCA (left column) as well as the sizes of the obtained sparsified sets $R'$ and $E'$ (middle and right columns, respectively). In parenthesis, we report the time it took to sparsifty each of the sets.
In Table \ref{tab:exp:stylegan_graph_time} we report the sizes of the vertex set $|\mathcal{G}|_\mathcal{V}$ and edge set $|\mathcal{G}|_\mathcal{E}$ of the resulting graph $\mathcal{G}$ build on the sparsified sets $R'$ and $E'$.

\begin{table}[h]
\caption{Results of the sparsification applied to $R$ and $E$ sets of different sizes. In the middle and right columns, we show the size of the obtained sparsified sets as well as the elapsed time in parenthesis.}
\label{tab:exp:stylegan_sparsification_time}
\vskip 0.15in
\begin{center}
\begin{small}
\begin{sc}
\begin{tabular}{c|c|c}
\toprule 
$|R|$, $|E|$ & $|R'|$ [time] & $|E'|$ [time] \\
 \midrule
 \midrule
 $50000$ & $9380$ [1h 32min] & $5349$ [54min] \\ 
 \midrule
 $10000$ & $2612$ [3min] & $1668$ [2min] \\ 
 \midrule
 $5000$ & $1387$ [50s] & $888$ [31s] \\ 
 \midrule
 $1000$ & $390$ [2s] & $295$ [2s] \\
\bottomrule
\end{tabular}
\end{sc}
\end{small}
\end{center}
\vskip -0.1in
\end{table}

\begin{table}[h]
\caption{Size of the vertex and edge sets obtained when building $\varepsilon$-graph on the sparsified sets $R'$ and $E'$ of sizes shown in Table~\ref{tab:exp:stylegan_sparsification_time}. The elapsed time when building the graph is shown in parenthesis.}
\label{tab:exp:stylegan_graph_time}
\vskip 0.15in
\begin{center}
\begin{small}
\begin{sc}
\begin{tabular}{ c|c|c|c } 
 \toprule
$|R|$, $|E|$ & $|R'|$ & $|E'|$ & $|\mathcal{G}|_\mathcal{V} + |\mathcal{G}|_\mathcal{E}$ [time]  \\
 \midrule
 \midrule
 $50000$ & $9380$ & $5349$ &$18734$ [11min] \\ 
 \midrule
 $10000$ & $2612$  & $1668$ & $5468$ [40s] \\ 
 \midrule
 $5000$ & $1387$ & $888$ & $2930 [11s]$ \\ 
 \midrule
 $1000$ & $390$ & $295$ & $863$ [1s]\\

\bottomrule
\end{tabular}
\end{sc}
\end{small}
\end{center}
\vskip -0.1in
\end{table}

\textbf{Varying $\varepsilon$ and $\delta$ parameters} 
In Figure~\ref{fig:stylegan-eps-delta}, we visualize $\mathcal{P}, \mathcal{R}, c(\mathcal{G})$ and $q(\mathcal{G})$ scores obtained when varying the sparsification parameter $\delta$ (left and middle), and distance threshold $\varepsilon$ (middle and right). These results were obtained on $R$ and $E$ sets of size $5000$. The middle panel corresponds to the parameters chosen in Section 5.%~\ref{sec:exp:gan}. 
We observe only slight changes in GeomCA scores when increasing $\varepsilon$ from $\varepsilon(10)$ to $\varepsilon(30)$. On the other hand, decreasing $\delta$ results in more significant changes that, however, still reflect the correct structure of $R$ and $E$. In particular, we observe that $\mathcal{P}, \mathcal{R}$ scores increase, while the network quality $q(\mathcal{G})$ decreases. This means that the connected components obtained when $\delta = 0.8 \cdot \varepsilon(10)$ contain more points than in case of $\delta = \varepsilon(10)$ but are also of lower quality. Note that the slight variations in network consistency are the result of applying sparsification with different parameters.

\subsection{VGG16 Model} \label{app:vgg16_model}

In Section 6,%~\ref{sec:exp:vgg16}, 
we defined two versions of the experiment where $R$ and $E$ sets contained $5$ different classes of ImageNet datasets each. In version $1$, we chose $R$ to contain representations of images of classes \emph{digital clock} (530), \emph{espresso maker} (550), \emph{frying pan} (557), \emph{mixing bowl} (659) and \emph{stove} (827), while $E$ contained \emph{Norwegian elkhound} (174), \emph{Weimaraner} (178) , \emph{Border terrier} (182), \emph{golden retriever} (207), \emph{Gordon setter} (214). In total, $R$ and $E$ contained $6258$ and  $6500$ representations, respectively. In version $2$, we randomly chose $R$ to contain \emph{Dungeness crab} (118), \emph{shopping basket} (791), \emph{lacewing} (318), \emph{ski} (795) and \emph{altar} (406), while $E$ contained \emph{fiddler crab} (120), \emph{sliding door} (799), \emph{sloth bear} (297), \emph{beagle} (162) and \emph{ladle} (618). In total, $R$ and $E$ each contained $6500$ representations. The $\varepsilon(10)$ threshold was estimated to $196.34$ and $197.59$ in version $1$ and $2$, respectively. Moreover, we used neighborhood size $k = 3$ for IPR and $L_0 = 64, \gamma = 1/128, i_{\max} = 100$ and $n = 1000$ for GS.
\end{document}